# Schrödinger Neural Network and Uncertainty Quantification: Quantum Machine

M. M. Hammad

Faculty of Science, Department of Mathematics and Computer Science, Damanhour University, Egypt
Email: m_hammad@sci.dmu.edu.eg
Orcid: https://orcid.org/0000-0003-0306-9719

## Abstract

We introduce the Schrödinger Neural Network (SNN), a principled architecture for conditional density estimation and uncertainty quantification inspired by quantum mechanics. The SNN maps each input to a normalized wave function on the output domain and computes predictive probabilities via the Born rule. The SNN departs from standard parametric likelihood heads by learning complex coefficients of a spectral expansion (e.g., Chebyshev polynomials) whose squared modulus yields the conditional density $p(y|x) = |\psi_x(y)|^2$ with analytic normalization. This representation confers three practical advantages: positivity and exact normalization by construction, native multimodality through interference among basis modes without explicit mixture bookkeeping, and yields closed-form (or efficiently computable) functionals—such as moments and several calibration diagnostics—as quadratic forms in coefficient space. We develop the statistical and computational foundations of the SNN, including (i) training by exact maximum-likelihood with unit-sphere coefficient parameterization, (ii) physics-inspired quadratic regularizers (kinetic and potential energies) motivated by uncertainty relations between localization and spectral complexity, (iii) scalable low-rank and separable extensions for multivariate outputs, (iv) operator-based extensions that represent observables, constraints, and weak labels as self-adjoint matrices acting on the amplitude space, and (v) a comprehensive framework for evaluating multimodal predictions. The SNN provides a coherent, tractable framework to elevate probabilistic prediction from point estimates to physically inspired amplitude-based distributions.

## Keywords

Schrödinger Neural Network, mixture density networks, normalizing flows, energy-based models, uncertainty quantification, physics informed neural networks, conditional density estimation, Born rule, spectral methods (Chebyshev).

## 1. Introduction

Predictive modelling in contemporary machine learning increasingly requires not merely point estimates but full, well-calibrated probability distributions that support risk-aware decision making, scientific inference, and robust operation under uncertainty; this requirement is particularly acute in domains where outputs are intrinsically multi-valued, where heteroscedasticity is pervasive, or where downstream actions are sensitive to tail behavior. Conventional neural-network heads that produce a single scalar (or a finite set of moments) [1-6] are fundamentally limited in these settings because they collapse rich conditional structure into a single summary statistic and thereby obscure important aspects of uncertainty and multi-modality; Gaussian heteroscedastic regressors extend this by permitting input-dependent variance but remain unimodal and symmetric, making them ill suited to inverse problems and ambiguous mappings in which the conditional distribution of the target given the input can exhibit multiple disjoint modes or pronounced skew. Mixture density networks (MDNs) [7] and their uses [8-16] address multimodality by parameterizing the conditional density as a finite mixture— commonly of Gaussian components— whose parameters depend on the input. While flexible and widely used, MDNs face practical challenges: the number of components must typically be chosen in advance, training can suffer from label-switching and degeneracy, and interpretability and stability degrade as component count grows. Normalizing flows (NFs) [17-26] model expressive conditional densities via invertible transforms from a simple base to a target distribution, offering exact likelihoods and great flexibility in high dimensions. However, invertibility and tractable-Jacobian constraints restrict architectures, often requiring deep, carefully engineered networks to capture complex multimodal structure. Energy-based models (EBMs) [27-32] provide a conceptually appealing unnormalized density framework but typically trade exact likelihoods for contrastive or sampling-based objectives that are computationally intensive and sensitive to sampler bias. Methods that target selective aspects of uncertainty—quantile regression, conformal prediction, or bootstrap-like schemes—offer robust guarantees for specific decision tasks but do not yield a generative statement about the conditional law and therefore cannot, without augmentation, deliver full probabilistic reports such as quantiles at arbitrary levels, analytic Bayes decisions, or closed-form risk integrals. Collectively, these limitations highlight a persistent tension: expressive conditional families that admit exact, tractable likelihoods and analytic downstream quantities are rare, and the practical remedies to patch limitations—mixture proliferation, variance floors, deep invertible stacks, sampler engineering—introduce their own engineering complexity and failure modes.

Motivated by this gap between expressive flexibility and analytic tractability, we propose a principled alternative: represent the conditional uncertainty associated with each input as a normalized wave function defined over the output domain and obtain probabilities by reading the squared modulus of that wave function according to Born's rule. This idea—drawing conceptual inspiration from de Broglie's association of waves with individual systems and from Born's statistical interpretation of amplitudes in quantum mechanics [33-35]—recasts conditional density estimation as the problem of estimating, for each input $\mathbf{x}$, a function $\psi_\mathbf{x}(\mathbf{y})$ in a Hilbert space such that $p(\mathbf{y}|\mathbf{x}) = |\psi_\mathbf{x}(\mathbf{y})|^2$. In this work, we introduce the Schrödinger neural network (SNN) as a novel approach that operationalizes this view. The SNN trains a neural network to compute the complex coefficients of a spectral expansion of $\psi_\mathbf{x}$ using an orthonormal basis of Chebyshev polynomials, ensuring numerical stability and efficiency. Two structural consequences of this design are central: first, positivity and normalization of the predictive density become structural properties of the representation rather than constraints to be enforced numerically—if the basis is orthonormal under the chosen measure then $\int|\psi_\mathbf{x}|^2$ reduces to the squared $\ell_2$-norm of the coefficient vector and unit normalization is achieved by a simple spherical projection of the network's coefficient output; second, multimodality and asymmetry emerge naturally via interference among basis modes in the amplitude whose squared modulus produces cross terms that can create separated peaks and complex shapes without explicit mixture components. These properties yield a compact and expressive family of conditional densities that can be trained by maximizing the exact log-likelihood under Born's rule, avoiding partition-function estimation, Jacobian determinants, and sampling-based surrogates. Beyond these pragmatic advantages, the wave–amplitude formulation affords a rich analytical calculus: moments, quantiles, cumulative distribution functions, and many diagnostics reduce to quadratic forms in coefficient space with precomputable matrices; spectral content and derivative-based quadratic forms give direct measures of smoothness and nodal complexity; and the formal analogy with Schrödinger operators motivates a family of variational regularizers (kinetic and potential energies) that have transparent statistical interpretations as penalties on spectral roughness and tail mass.

The SNN therefore synthesizes the best features of several approaches while avoiding many of their pitfalls: like mixture models and flows it is capable of representing complex, multimodal conditionals; like NFs it affords exact likelihood evaluation for training; unlike MDNs it does not require an a priori component count or suffer from label switching among components because multi-modality is an emergent property of amplitude interference rather than an indexable set of components; unlike many EBMs it does not demand sampling to evaluate the likelihood; and unlike quantile or conformal heads it produces a full generative density from which any decision-theoretic quantity can be computed analytically or with high-accuracy quadrature. At the same time, the SNN's spectral parameterization gives practitioners explicit knobs —basis order, compactification mapping, kinetic weight, and phase allowance—that have clear approximation-theoretic and numerical interpretations, enabling systematic capacity control and diagnostics rather than ad hoc tuning. Equally important, the operator perspective implicit in the wave formalism allows natural and computationally cheap encodings of constraints and weak supervision: multiplication operators represent expectations of arbitrary functions of the output (moments, tail indicators, safety constraints), projection operators encode interval or censored observations, and positive operator-valued measures capture noisy or coarsened labels—all of which reduce to sparse matrix operations in the spectral basis and are therefore straightforward to incorporate into likelihood-based training.

Across the history of the physical sciences, progress has often hinged on reframing the basic unit of analysis. Classical mechanics treated bodies as point particles following deterministic trajectories, and classical fields as continuous entities propagating through space. Within this worldview, measurement was a passive act that revealed preexisting properties, uncertainty was a reflection of incomplete information, and causality unfolded along predictable lines. The advent of quantum mechanics disrupted this foundation by replacing definite trajectories with probability amplitudes, supplanting naive realism with a more operational account of measurement, and elevating structure—symmetry, operators, and spectra—over intuition. Matter ceased to be a collection of tiny billiard balls and became an entity whose behavior is encoded in complex-valued amplitudes, whose probabilities arise from squared magnitudes, and whose measurable properties are defined by actions of operators. In the present work, a parallel reorientation is underway in the form of amplitude-first modeling with SNNs. Here, the basic unit is no longer a labeled pair that binds one input to one definitive output, but a conditional amplitude over possible outputs whose squared magnitude is the predictive density and whose phase carries functional content through interference. As quantum mechanics changed how physicists think about matter, we envision that amplitude-based learning can redefine the perspective of data scientists on a data point, elevating it from a single observation to a structured constraint within a complex representation.

This paper articulates the SNN architecture, justifies its design both mathematically and practically, and delineates a range of extensions and diagnostics that together form a coherent framework for uncertainty-aware prediction. Our contributions are as follows. First, we introduce the SNN as an architecture that maps inputs to normalized spectral amplitudes and defines conditional densities via Born's rule; we present the concrete spectral synthesis in a Chebyshev basis, the unit-sphere parameterization of coefficients that enforces analytic normalization, and the associated negative log-likelihood (NLL) training objective. Second, we study representational properties and approximation rates: we relate the SNN's expressive capacity to the density of polynomial spans in $L^2$ and to the neural-network universal approximation of the coefficient map, and we characterize how smoothness and analyticity of conditional densities affect spectral convergence, providing practical guidance for choosing basis order and compactification. Third, we develop a family of physics-inspired quadratic regularizers—kinetic energy to penalize spectral roughness and soft confining potentials to control tail mass—and show how these terms are precomputable quadratic forms in coefficient space that integrate seamlessly with gradient-based optimization. Fourth, we extend the formalism to an operator calculus that encodes observables, constraints, and weak supervision as self-adjoint matrices with efficient quadratic evaluations. Finally, we provide empirical demonstrations and ablations that illustrate the SNN's behavior on representative multimodal inverse problems and heteroscedastic regression benchmarks, examine calibration and out-of-distribution diagnostics rooted in amplitude geometry (entropy, kinetic energy, nodal stability).

The remainder of the paper is organized as follows. Section 2 surveys background and related work on conditional density estimation, positioning our approach with respect to MDNs, NFs, and EBMs. Section 3 articulates two foundational principles from quantum mechanics and motivates an amplitude-first learning paradigm that underpins our methodology. Section 4 presents the formal definition of the SNN, including its probabilistic semantics and training objective. Section 5 describes the network architecture, training procedure, and hyperparameter selection, with an emphasis on reproducibility and practical implementation details. Section 6 examines the role of complex coefficients, clarifies gauge invariance, and demonstrates how interference arises and can be modulated within the learned representation. Section 7 connects the operator postulate to the SNN by defining observables and measurement rules consistent with the model's amplitude-based formulation. Section 8 introduces kinetic and potential quadratic regularizers, interprets their influence on uncertainty and smoothness, and provides guidance for their use in practice. Section 9 proposes a comprehensive framework for evaluating multimodal predictions and verifying SNN outputs against baselines. Section 10 generalizes the SNN framework to handle multivariate outputs. Section 11 concludes by summarizing contributions, discussing limitations, and outlining opportunities for future research.

## 2. Background and Related Work: Neural Approaches to Conditional Density Estimation

This section reviews key methods motivating our wave-function–based conditional density estimation, surveying neural estimators—MDNs, NFs, and EBMs—while emphasizing expressiveness, training stability, and computational trade-offs.

MDNs parameterize $p(\mathbf{y}|\mathbf{x})$ as a finite mixture whose weights and component parameters are functions of the input $\mathbf{x}$. In the most common form with $M$ Gaussian components [7],

$$p(\mathbf{y}|\mathbf{x}) = \sum_{m=1}^{M} \pi_m(\mathbf{x}) \mathcal{N}(\mathbf{y}; \boldsymbol{\mu}_m(\mathbf{x}), \Sigma_m(\mathbf{x})), \tag{1}$$

where $\{\pi_m\}$ is a set of weights such that $\pi_m(\mathbf{x}) \geq 0$, $\sum_{m=1}^{M} \pi_m(\mathbf{x}) = 1$, $\mathcal{N}(\mathbf{y}; \mu_m(\mathbf{x}), \Sigma_m(\mathbf{x}))$ is the Gaussian probability density function (PDF) for the $m$-th component with mean $\boldsymbol{\mu}_m(\mathbf{x})$ and variance $\Sigma_m(\mathbf{x})$. This design is attractive because it directly encodes multimodality and heteroscedasticity: by varying $\boldsymbol{\mu}_m(\mathbf{x})$, $\Sigma_m(\mathbf{x})$ with $\mathbf{x}$, the conditional law can split into multiple lobes or inflate/deflate as the covariates change. Training proceeds by maximizing the exact conditional log-likelihood, and inference is straightforward. However, the approach introduces several practical issues. First, component selection is architectural: one must choose $M$ a priori or resort to heuristics that grow or prune components; misspecification here leads to either underfitting (too few modes) or brittle training (too many degrees of freedom). Second, responsibility degeneracy is common: without careful initialization and variance floors, one or a few components dominate, with others collapsing to negligible weights or to singular covariances that spike the likelihood on a small set of points. Third, label switching—the permutation symmetry among components—complicates interpretability and monitoring, as component identities shift across training and between batches. Fourth, scaling to higher output dimensions requires either diagonal covariances (sacrificing correlation structure) or full covariances with cubic costs per component; structured covariances alleviate but do not eliminate this trade-off. Finally, although MDNs often yield competitive calibration, their probabilities inherit the tail behavior and symmetry of the chosen component family, which can be limiting when true conditionals are skewed or heavy-tailed in ways not well approximated by a small Gaussian mixture.

NFs construct flexible densities by learning an invertible map $T_\mathbf{x}: \mathbb{R}^d \to \mathbb{R}^d$ that pushes a simple base distribution $p_Z$ (e.g., standard Gaussian) to the target conditional $p(\cdot|\mathbf{x})$. The change-of-variables formula gives [20]

$$p(\mathbf{y}|\mathbf{x}) = p_Z(T_\mathbf{x}^{-1}(\mathbf{y})) |\det \nabla_\mathbf{y} T_\mathbf{x}^{-1}(\mathbf{y})|. \tag{2}$$

Architectures (coupling layers, autoregressive transforms, continuous-time flows) are designed so that both $T_\mathbf{x}^{-1}$ and the log-determinant are tractable, enabling exact maximum-likelihood training with stable gradients. Flows are highly expressive, scale well to moderate and high output dimensions, and offer fast sampling. Their principal limitations arise from invertibility and Jacobian constraints, which restrict the family of admissible transformations and can make it difficult to capture sharply separated modes without deep stacks or carefully engineered couplings. In conditional settings, the dependence of $T_\mathbf{x}$ on $\mathbf{x}$ must be modeled by a conditioner network that can be brittle if the mapping from inputs to flow parameters is highly non-linear.

Moreover, several studies have documented counterintuitive likelihood behavior under distribution shift—e.g., high likelihood assigned to out-of-distribution inputs—unless the architecture and regularization are carefully tuned. While not universal, these behaviors highlight that exact likelihood alone does not guarantee calibrated uncertainty in finite-sample.

An EBM specifies a conditional density over outputs **y** given inputs **x** using a learnable energy function $E_\theta(\mathbf{y}, \mathbf{x})$ [32]:

$$p(\mathbf{y}|\mathbf{x}) = \frac{\exp(-E_\theta(\mathbf{y}, \mathbf{x}))}{Z_\theta(\mathbf{x})}, \qquad Z_\theta(\mathbf{x}) = \int \exp(-E_\theta(\mathbf{y}, \mathbf{x})) \, d\mathbf{y}, \tag{3}$$

where $Z_\theta(\mathbf{x})$ is the partition function (normalizer). It depends on **x** and **θ**, and is usually intractable to compute exactly in continuous, high-dimensional spaces. This formulation is extremely flexible—any positive density can be represented in this way— but it shifts the difficulty to estimating the partition function. In principle, maximum likelihood requires gradients of $\log Z_\theta(\mathbf{x})$, and one has

$$\nabla_\theta \log Z_\theta(\mathbf{x}) = \mathbb{E}_{p_\theta(\cdot|\mathbf{x})}[\nabla_\theta E_\theta(\mathbf{y}, \mathbf{x})], \tag{4}$$

i.e., an expectation with respect to the model's own conditional distribution. Consequently, computing the required gradients (or even approximating) entails drawing samples from the very model being optimized, creating a circular dependence between likelihood optimization and model-based sampling that complicates training in practice. Training therefore relies on contrastive surrogates (e.g., noise-contrastive estimation) or Markov chain Monte Carlo to generate negative samples, with the attendant computational cost and sensitivity to sampler bias and mixing. Recent amortized or score-matching variants mitigate some costs, but exact likelihoods are generally unavailable; this complicates model selection on proper scoring rules and can impair probability calibration. EBMs are strong when one needs to encode hard constraints (energies that diverge outside feasible sets) or when unnormalized potentials align with domain knowledge, but practical deployment often requires significant engineering to stabilize training and to produce reliable uncertainties.

Across all three families, MDNs, NFs, and EBMs, enforcing positivity and normalization is either a parametric constraint (MDNs), a Jacobian calculation (NFs), or an intractable partition function (EBMs). Moreover, many downstream decision quantities—moments, quantiles, tail risks—require either numerical integration or additional modeling heads. These observations motivate a representation in which (i) positivity and normalization are structural, not numerical constraints, (ii) multimodality arises natively without component bookkeeping, and (iii) downstream functionals are analytic in the model parameters. The amplitude-based view adopted later in the paper aims precisely at this combination.

## 3. Two Quantum Mechanics Pillars: Amplitude-First Learning

This section introduces the conceptual foundation of the SNN and underscores the importance of these principles for statistical learning. The formulation details—how amplitudes are parameterized, how normalization is enforced, and which losses and regularizers realize the Born-consistent objective—are deferred to the next sections, where the SNN is specified precisely.

The organizing idea behind an SNN is that statistical prediction can be reformulated as wavefunction estimation: for each input **x**, the model associates a complex amplitude field $\psi_\mathbf{x}$ defined over the outcome domain, and the predictive law arises by passing from amplitudes to probabilities through the squared modulus. Two pillars of quantum theory give this formulation both meaning and discipline. First, de Broglie's association of waves with individual physical systems provides the conceptual license to attach a wave to each datum: just as a particle with momentum **p** is accompanied by a phase factor $e^{i\mathbf{k}\cdot\mathbf{r}}$ with $\mathbf{k} = \mathbf{p}/\hbar$, a data case **x** is accompanied by a structured amplitude field $\psi_\mathbf{x}(.)$ defined on the outcome domain. Second, Max Born's statistical reading of wave amplitudes provides the operational rule that turns these complex fields into empirical statements: the predictive law for an outcome **y** conditional on **x** is $p(\mathbf{y}|\mathbf{x}) = |\psi_\mathbf{x}(\mathbf{y})|^2$ (with appropriate normalization). In the SNN, the neural network plays the role of a parametric dynamical law that determines $\psi_\mathbf{x}$ from **x**, in exact analogy to the way Schrödinger's equation generates $\psi$ from a Hamiltonian.

De Broglie's pillar is a representational claim: waves serve as function spaces rich enough to capture locality, structure, and interference. At its core, this representational pillar starts from a simple observation: superposition turns waves into adaptable objects. In physical systems, a single plane wave is delocalized, but superpositions of waves build localized packets whose group properties mirror classical kinematics. Translating that idea, an SNN rejects the notion that a datum must be mapped to a single number (a point estimate) or even to a small fixed family of numbers (e.g., the mean and variance of a Gaussian). Instead, the datum is mapped to a function $\psi_\mathbf{x}$, which can be thought of as a "packet" whose shape, width, oscillation, and nodal structure encode everything one wishes to predict about **y**|**x**. In de Broglie's original language, interference, dispersion, and phase are fundamental mechanisms by which the wave encodes detailed behavior. In learning, interference corresponds to the presence of cross-terms in $|\psi_\mathbf{x}|^2$ that can produce multi-modality, skewness, and heavy tails without resorting to explicit mixture carpentry. Dispersion corresponds to how smoothness (or its controlled violation) in $\psi_\mathbf{x}$ translates to the sharpness and regularity of the predicted density. Phase, though unobservable at the level of $|\psi_\mathbf{x}|^2$, controls constructive and destructive interference among basis components—effects unattainable by convex combinations of classical probability components— and thus acts as a hidden degree of freedom through which the model adjusts shape while preserving probability mass. By embracing a wave for each input, the SNN elevates representation from "numbers about **y**" to "a field over **y**", thereby restoring to learning the same expressive geometry that made de Broglie's wave viable in physics. Amplitude-based models are therefore natural when the underlying generative process exhibits coherent interference, whereas direct probabilistic models such as MDNs, NFs, or autoregressive architectures suffice when interference is absent.

On the other hand, Born's pillar provides the operational semantics. The Born rule states that the physically meaningful content of a quantum state is probabilistic and becomes manifest only when one passes from complex amplitudes to measurable likelihoods. Its core claim is not a technical convenience but a foundational prescription that reconciles linear superposition with empirical observation. Interference, for example, arises because amplitudes combine additively before probabilities are computed; the Born rule translates that linear structure into frequencies of outcomes that match experimental statistics. In this way, it bridges the abstract Hilbert-space description and the laboratory, ensuring that predictions are both internally consistent and operationally testable.

For statistical learning, from a technical perspective, the Born quadratic law is decisive because it simultaneously guarantees nonnegativity and supports exact normalization in orthonormal bases. This analytic normalization has two statistical dividends. First, it ensures that training under the NLL is a strictly proper scoring rule. Second, it removes an entire class of computational pathologies—drift in normalization constants, blow-ups at the tails, or implicit competition between scale and shape parameters—that beset generic density heads.

The SNN architecture enforces the same separation between representation and observation that underpins quantum mechanics: amplitude space is where composition and learning occur, whereas probability space is where decisions and evaluations are made. The benefit is twofold. First, uncertainty is native, not an afterthought. The model returns a full predictive distribution that can capture multi-modal structure, heavy tails, and heteroscedastic spread. Second, expressive interactions among basis functions happen at the amplitude level, allowing rich probability landscapes to emerge after the non-linear mapping prescribed by the Born rule.

Notably, in quantum mechanics, the wavefunction itself is not directly observable, yet it compactly encodes all knowable information about a system—position, momentum, energy, spin—so that measurable statistics are recovered only after applying the Born rule. Analogously, a single learned object $\psi_\mathbf{x}(\cdot)$ serves as a unified complex-valued representation generated for each input, encoding all information relevant to the conditional distribution of **y**|**x**: point forecasts, conditional dispersion, skewness, multimodality, tail behavior, and input-dependent higher-order functionals (quantiles, risk measures) are all read from the same field via the Born map $|\psi_\mathbf{x}(\mathbf{y})|^2$. The SNN places a single mathematical object—the conditional wavefunction $\psi_\mathbf{x}(\mathbf{y})$—at the center of learning and decision making, and from this construct recovers, within one model, the behaviors of many seemingly disparate machine-learning paradigms. Predictive tasks that are traditionally solved by distinct model families—point regression, probabilistic regression, classification, quantile estimation, density modeling, energy-based scoring, and even structured prediction—emerge as alternate ways of querying or shaping the same conditional density. The unifying role of the wavefunction is to supply an expressive, differentiable surrogate for all target-side uncertainty; every supervision signal and every decision rule becomes a functional acting on $\psi_\mathbf{x}$ rather than a bespoke architecture. This viewpoint eliminates the proliferation of task-specific heads and training criteria and replaces them with a single, coherent object whose amplitude governs probabilities while its phase can encode useful auxiliary structure such as continuity constraints, inductive biases, or latent fields tied to the geometry of the output space.

## 4. Schrödinger Neural Network: Mathematical Formulation

In the SNN, a standard feed-forward neural network does not predict a target directly; instead, it generates coefficients of a spectral expansion for a complex wave function $\psi$ defined on the output space. The wave function is represented as a finite linear combination of Chebyshev polynomials of the first kind, which exploit orthogonality to obtain rapid convergence for smooth targets, evaluated on a suitably transformed output coordinate.

To formalize the setting, consider supervised pairs $\{(\mathbf{x}_i, y_i)\}_{i=1}^N$ with features $\mathbf{x} \in \mathcal{X} \subset \mathbb{R}^d$ and a scalar output $y$ taking values in a bounded interval $\mathcal{Y} = [a, b]$. We first present the one-dimensional construction and subsequently indicate how it extends to multivariate outputs. The SNN associates to each conditioning input **x** a complex-valued amplitude (or "wave function") $\psi_\mathbf{x}: \mathcal{Y} \to \mathbb{C}$ that is represented spectrally in a finite, numerically well-conditioned Chebyshev basis. Let $f: [a, b] \to [-1,1]$ be the linear bijection

$$\xi = f(y) = \frac{2(y-a)}{b-a} - 1, \tag{5}$$

that relocates the physical output domain $\mathcal{Y}$ to the canonical Chebyshev interval. On $[-1,1]$ the $k$-th Chebyshev polynomial of the first kind is $T_k(\xi) = \cos(k \cos^{-1} \xi)$, and the family $\{T_k\}_{k \geq 0}$ is orthogonal with respect to the weight $w(\xi) = (1 - \xi^2)^{-1/2}$. Specifically,

$$\int_{-1}^{1} T_j(\xi)T_k(\xi)w(\xi)d\xi = \begin{cases} \pi, & j = k = 0, \\ \pi/2, & j = k \geq 1, \\ 0, & j \neq k, \end{cases} \tag{6}$$

so that the re-scaled system

$$\varphi_k(\xi) = \kappa_k T_k(\xi), \tag{7}$$

with $\kappa_0 = \pi^{-1/2}$ and $\kappa_k = (2/\pi)^{1/2}$ for $k \geq 1$, is orthonormal in $L^2([-1,1], d\mu)$, $d\mu(\xi) = w(\xi)d\xi$, i.e.,

$$\int_{-1}^{1} \varphi_j(\xi)\varphi_k(\xi)w(\xi)d\xi = \delta_{jk}. \tag{8}$$

For each conditioning input $\mathbf{x}$, the feed-forward neural network outputs a coefficient vector

$$\mathbf{c}(\mathbf{x}) = (c_0(\mathbf{x}), \ldots, c_K(\mathbf{x})), \tag{9}$$

and the wave function (in $\xi$-space) is the degree-$K$ spectral expansion

$$\psi_{\mathbf{x}}(\xi) = \sum_{k=0}^{K} c_k(\mathbf{x})\varphi_k(\xi). \tag{10}$$

Equivalently, in $y$-coordinates,

$$\psi_{\mathbf{x}}(y) \equiv \psi_{\mathbf{x}}(f(y)) = \sum_{k=0}^{K} c_k(\mathbf{x})\varphi_k(f(y)). \tag{11}$$

Because $\{\varphi_k\}$ is orthonormal in $L^2([-1,1], d\mu)$ with $d\mu(\xi) = w(\xi)d\xi$, normalization under the Born prescription becomes analytic:

$$\int_{-1}^{1} |\psi_x(\xi)|^2 d\mu(\xi) = \sum_{k=0}^{K} |c_k(\mathbf{x})|^2 \equiv \|\mathbf{c}(\mathbf{x})\|_2^2. \tag{12}$$

Consequently, the conditional density with respect to $d\mu$ is

$$p(\xi|\mathbf{x}) = \frac{|\psi_{\mathbf{x}}(\xi)|^2}{\|\mathbf{c}(\mathbf{x})\|_2^2}. \tag{13}$$

To express the density with respect to the Lebesgue measure on $\mathcal{Y}$, note that $d\mu(\xi) = w(\xi)d\xi = w(f(y))f'(y)dy$. A change of variables yields

$$p(y|\mathbf{x}) = \frac{|\psi_{\mathbf{x}}(f(y))|^2 w(f(y))f'(y)}{\int_a^b |\psi_{\mathbf{x}}(f(u))|^2 w(f(u))f'(u)du} = \frac{|\psi_{\mathbf{x}}(f(y))|^2 w(f(y))f'(y)}{\|\mathbf{c}(\mathbf{x})\|_2^2}. \tag{14}$$

It is often convenient to absorb the measure factor into a rescaled amplitude

$$\Psi_{\mathbf{x}}(y) = \psi_{\mathbf{x}}(f(y))\sqrt{w(f(y))f'(y)}, \quad \int_a^b |\Psi_{\mathbf{x}}(y)|^2 dy = \|\mathbf{c}(\mathbf{x})\|_2^2, \tag{15}$$

so that $p(y|\mathbf{x}) = |\Psi_{\mathbf{x}}(y)|^2/\|\mathbf{c}(\mathbf{x})\|_2^2$ is a probability density over $[a,b]$ with respect to $dy$. In particular, if we enforce $\|\mathbf{c}(\mathbf{x})\|_2 = 1$ for all $\mathbf{x}$ by projecting the network output $\mathbf{z}(\mathbf{x})$ onto the unit sphere

$$\mathbf{c}(\mathbf{x}) = \frac{\mathbf{z}(\mathbf{x})}{\|\mathbf{z}(\mathbf{x})\|_2}, \quad \int_{-1}^{1} |\psi_{\mathbf{x}}(\xi)|^2 d\mu(\xi) = 1, \tag{16}$$

holds identically. This normalization is analytic, does not depend on numerical quadrature, and is stable under gradient-based training because it reduces to an explicit algebraic rescaling of the coefficient vector.

More generally, for an arbitrary real basis $\{\varphi_k\}$ and inner product $\langle u, v \rangle = \int uv\, d\nu$ (with $d\nu$ either $dy$ or $d\mu$), normalization is the quadratic form

$$\|\psi_{\mathbf{x}}\|_{L^2(d\nu)}^2 = \int \left|\sum_k c_k(\mathbf{x})\varphi_k\right|^2 d\nu = \mathbf{c}^H(\mathbf{x})G\mathbf{c}(\mathbf{x}), \tag{17.1}$$

where $\mathbf{c}^H$ is the Hermitian (conjugate) transpose of $\mathbf{c}$. Gram matrix $G \in \mathbb{R}^{(K+1)\times(K+1)}$ has entries

$$G_{jk} = \int \varphi_j \varphi_k d\nu. \tag{17.2}$$

In this setting, a robust "unit-sphere" projection is

$$\mathbf{c}(\mathbf{x}) = \frac{\mathbf{z}(\mathbf{x})}{\sqrt{\mathbf{z}^H(\mathbf{x})G\mathbf{z}(\mathbf{x}) + \varepsilon}}, \tag{18}$$

which enforces $\mathbf{c}^H G\mathbf{c} = 1$ analytically (and reduces to $\|\mathbf{c}\|_2 = 1$ when $G = I$). Here, $\varepsilon > 0$ is a small constant that prevents division overflow early in training.

Given a dataset $\{(\mathbf{x}_i, y_i)\}_{i=1}^{N}$, the core training objective is the negative conditional log-likelihood, from (14),

$$\mathcal{L}_{NLL}(\theta) = -\frac{1}{N}\sum_{i=1}^{N} \log p_\theta(y_i|\mathbf{x}_i) = -\frac{1}{N}\sum_{i=1}^{N} \log|\psi_{\mathbf{x}_i}(\xi_i)|^2 + \log w(\xi_i) + \log f'(y_i) - \log\|\mathbf{c}(\mathbf{x})\|_2^2, \tag{19}$$

where $\theta$ denotes all network parameters that determine the coefficient map $\mathbf{x} \mapsto \mathbf{c}(\mathbf{x}) \equiv (c_0(\mathbf{x}), \ldots, c_K(\mathbf{x}))$ and $\xi_i = f(y_i)$. If normalization is enforced ($\|\mathbf{c}(\mathbf{x})\|_2^2 = 1$ or $\mathbf{c}^H(\mathbf{x})G\mathbf{c}(\mathbf{x}) = 1$), the last term vanishes, and the $\log w(\xi_i) + \log f'(y_i)$ terms are constant with respect to $\theta$; dropping constants yields the simple, well-conditioned objective

$$\mathcal{L}_{NLL}(\theta) = -\frac{1}{N}\sum_{i=1}^{N} \log |\psi_{\mathbf{x}_i}(\xi_i)|^2 = -\frac{1}{N}\sum_{i=1}^{N} \log \left|\sum_k c_k(\mathbf{x}_i)\varphi_k(\xi_i)\right|^2. \tag{20}$$

The objective depends only on $|\psi_\mathbf{x}|$ and is therefore invariant under global phase transformation $\psi \mapsto e^{i\phi}\psi$; the learning dynamics should respect this gauge symmetry to avoid unidentifiable directions in parameter space.

Representational expressiveness stems from two coupled universal approximation properties: (i) Chebyshev polynomials span a dense subspace of $L^2([-1,1], d\mu)$, so any square-integrable amplitude function $\psi(\cdot)$ can be approximated arbitrarily well in norm by $\sum_{k=0}^{K} c_k \varphi_k(\cdot)$ as $K \to \infty$; and (ii) feed-forward neural networks are universal approximators of continuous functions on compact subsets of $\mathbb{R}^d$, so the map $\mathbf{x} \mapsto \mathbf{c}(\mathbf{x})$ can approximate any continuous coefficient field. Therefore, under mild regularity, the SNN can approximate any continuous conditional square-root density $\mathbf{x} \mapsto \psi_\mathbf{x}(\cdot)$ in $L^2$, and hence any conditional density $p(y|\mathbf{x})$ via $p = |\psi|^2$. A finite truncation $K$ constrains the model's expressiveness. However, Chebyshev spectral methods are spectrally accurate for smooth targets: when the conditional density is sufficiently smooth (in practice: infinitely differentiable and without nearby singular behavior), the truncation error decreases exponentially fast in $K$; if the density is only finitely smooth, convergence is slower (algebraic). Non-smooth or highly multimodal densities require larger $K$.

Two implementation routes follow. The first evaluates the norm by accurate quadrature on a fixed grid, which is simple and basis-agnostic. The second enforces normalization exactly in coefficient space via the quadratic form $\mathbf{c}^*G\mathbf{c} = 1$, thereby removing one numerical source of training instability and sharpening the probabilistic interpretation of the amplitude. While either normalization strategy is valid, enforcing $\mathbf{c}^H G\mathbf{c} = 1$ by construction has three practical advantages for SNNs. First, it isolates model capacity and regularization effects to shape rather than scale, so that improvements in likelihood cannot be obtained by trivial rescaling of the amplitude. Second, it aligns training with the natural geometry of the parameter manifold: the coefficients live on the complex unit sphere modulo a global U(1) phase, and respecting that constraint eliminates flat directions associated with global phase changes. Third, it facilitates analytic and efficient computation of quadratic regularizers (discussed below) because their expectation values reduce to Hermitian forms $\mathbf{c}^H \mathbf{H}\mathbf{c}$ with fixed positive semidefinite operators $\mathbf{H}$. In all cases, numerical stabilizers—an additive $\varepsilon$ inside the logarithm and high-precision accumulation in quadrature when the grid is used—guard against overflow and underflow when amplitudes become sharply localized.

The probabilistic semantics confer immediate benefits for uncertainty quantification and decision-making. Given $p(y|\mathbf{x})$, one can compute posterior means, variances, quantiles, credible intervals, and risk-aware functionals (e.g., expected shortfall) by analytic or high-accuracy quadrature over the basis.

## 5. Simple Inverse Problems: SNN Architecture and Training Procedure

To illustrate the practical use of the SNN, we begin with a canonical one-dimensional inverse problem in which an observed variable is generated by a smooth, nonlinear forward process with small stochastic perturbations, and the task is to infer the latent variable that produced each observation. This setting is deliberately simple yet diagnostically rich: the forward map is locally non-invertible, so the inverse relation from observation to latent variable is many-to-one and therefore multimodal. Conventional point estimators and mean-squared-error regression collapse this structure into a single value and systematically misrepresent uncertainty, while MDNs can approximate multiple peaks only by increasing the number of components and still tend to over-fill the low-density valleys between modes. The SNN tackles the problem in an amplitude-first manner by learning a complex wave-amplitude over the latent domain whose squared modulus defines a properly normalized conditional density. We train the SNN by maximizing conditional log-likelihood on simulated pairs drawn from the forward process with additive noise, using smooth spectral bases for the amplitude and quadratic regularizers to control roughness and tail mass.

Concretely, we consider the forward mapping from a latent variable $t$ to an observation $x$ defined by
$$x = t + 0.30 \sin(2\pi t) + \epsilon, \tag{21}$$
where $\epsilon$ is uniformly distributed on $[-0.1, 0.1]$. In the absence of the noise term, the mapping is single-valued—each $t$ deterministically yields a unique $x$—but it is not monotone, and thus the inverse relation from $x$ to $t$ is globally many-to-one. The added bounded noise thickens the forward curve into a ribbon without altering this fundamental non-invertibility, so the correct posterior over $t$ given $x$ is frequently multimodal with peaks near the preimages of $x$ under the noiseless map.

Within the SNN, the mapping from the observation variable $x$ to the expansion coefficients $\mathbf{c}(x)$ is implemented by a standard multi-layer perceptron (MLP), selected for its established ability to approximate complex, nonlinear functions. The network receives a scalar input $x \in \mathbb{R}$ and propagates it through a sequential feed-forward stack comprising three fully connected hidden layers with 256 units each. Nonlinearity in the hidden layers is provided by the Gaussian error linear unit (GELU) [36], whose smooth, input-dependent gating is well suited to stable gradient-based optimization in deep architectures.

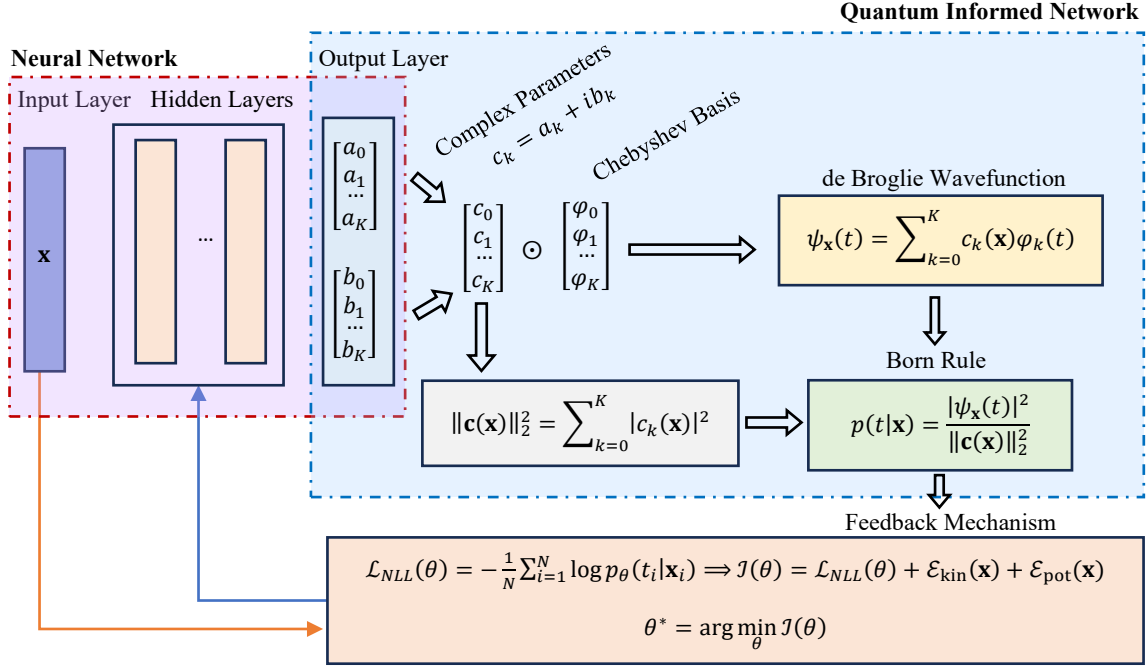

**Figure 1.** SNN architecture. The model maps an input $\mathbf{x}$ to a complex coefficient vector $\mathbf{c}(\mathbf{x})$ via a multilayer perceptron, splits real and imaginary channels, and expands over an orthonormal spectral basis $\varphi_k$ to form a complex amplitude $\psi_{\mathbf{x}}(t) = \sum_{k=0}^{K} c_k(\mathbf{x})\varphi_k(t)$. Born normalization enforces a valid conditional density $p(t|\mathbf{x}) = |\psi_{\mathbf{x}}(t)|^2/\|\mathbf{c}(\mathbf{x})\|_2^2$, implemented analytically from the coefficient norm to decouple scale from shape. Optional physics-inspired quadratic regularizers—kinetic (spectral smoothness) $\mathcal{E}_{\text{kin}}(\mathbf{x})$, (39), and potential (soft support control) $\mathcal{E}_{\text{pot}}(\mathbf{x})$, (40)—act directly in coefficient space. Training minimizes conditional negative log-likelihood; inference yields calibrated multimodal posteriors, point summaries (mean, median, modes), and uncertainty diagnostics, while complex phases enable interference that shapes valley depth between peaks without additional mixture components.

This configuration was chosen to afford adequate representational capacity for learning the inverse relation in (21), which is encoded implicitly in the spectral coefficient vector $\mathbf{c}(x)$. A schematic of the resulting architecture and data flow is presented in figure 1.

The final layer, which constitutes the output of the coefficient network, is a linear dense layer with a dimension directly tied to the basis expansion order. With a Chebyshev polynomial order of $K = 24$, the spectral expansion requires $K + 1 = 25$ coefficients. As the target SNN model is designed to operate with complex coefficients, $\mathbf{c}(x) \in \mathbb{C}^{25}$, the output layer must regress both the real and imaginary parts of each coefficient. Consequently, the output layer possesses $2(K + 1) = 50$ units, with the first 25 units representing the real parts ($\text{Re}[\mathbf{c}]$) and the subsequent 25 units representing the imaginary parts ($\text{Im}[\mathbf{c}]$). This distinct output configuration is the sole architectural feature differentiating the complex SNN (CSNN) from its real SNN (RSNN), and it is critically essential for exploiting the phase freedom in the complex space, which dramatically enhances the model's ability to accurately represent multi-modal and non-Gaussian conditional PDFs $p(t|x)$.

The optimization of the SNN parameters is achieved by minimizing the analytically Born-normalized NLL objective function using the ADAM optimizer. Regularization techniques are also integrated into the training objective, specifically, an $L_2$ regularization term on the coefficients, $\mathcal{L}_{\text{train}} = NLL_{\text{train}} + \lambda \|\mathbf{c}(x)\|^2$, to mitigate overfitting and promote generalization to unseen data. Hyperparameter selection, including the initial learning rate (set at $LR = 10^{-3}$) and the regularization coefficient ($\lambda = 10^{-5}$) was performed through fine-tuning based on the performance observed on a dedicated validation set. The training procedure was governed by an early stopping criterion.

The six panels of figure 2 provide a coherent assessment of an amplitude-first approach to multimodal inverse inference. Figure 2(a) shows stable optimization of the NLL augmented with a small coefficient penalty. The validation NLL typically attains a clear minimum at epoch 139, indicating an appropriate balance between fidelity to the data and spectral smoothness; the gap between training and validation NLLs serves as a direct check on generalization. Because the model enforces normalization through $\int |\psi|^2 dt$, the objective is a proper conditional likelihood and is directly comparable across epochs. Figure 2(b) visualizes the learned posterior field $p(t|x)$ and demonstrates recovery of the many-to-one inverse geometry created by the sinusoidal perturbation in the forward map. Multiple high-density ridges at fixed $x$ correspond to distinct plausible latent values $t$, while the overlaid (thinned) training pairs $(x, t)$ populate these ridges rather than the intervening valleys. This pattern indicates that the SNN allocates probability mass conservatively in low-support regions, avoiding the "over-fill" behavior often observed in mixture regressions with insufficient component control. Figure 2(c) examines $\psi(t|x^\star)$ at a representative input, $x^\star$.

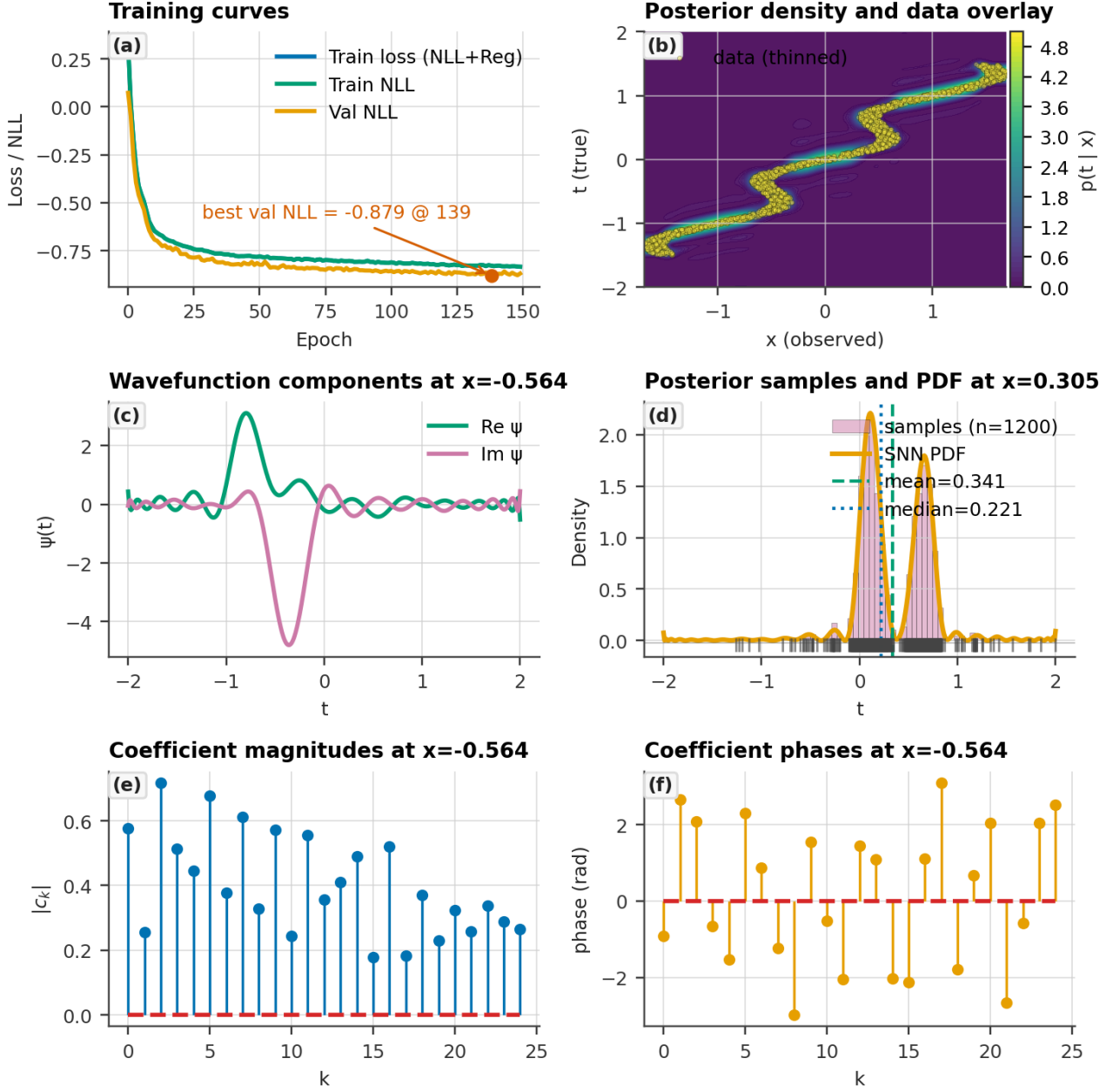

**Figure 2.** SNN for a non-invertible inverse map (21). The SNN models the conditional density $p(t|x)$ induced by the forward process $x = t + 0.30 \sin(2\pi t) + \epsilon$, $\epsilon \in U[-0.1, 0.1]$, over $t \in [-2,2]$ using a complex-amplitude expansion $\psi_\mathbf{x}(t) = \sum_{k=0}^{K} c_k(\mathbf{x}) \varphi_k(t)$ with orthonormal Chebyshev functions ($K = 24$). Training minimizes conditional negative log-likelihood with $\ell_2$ coefficient regularization $\lambda = 10^{-5}$ using Adam (learning rate $10^{-3}$, batches train/validation 128/256, 150 epochs). $p(t|x) = |\psi|^2 / \int |\psi|^2 dt$ is computed by trapezoidal quadrature. (a) Training curves: total loss (NLL+regularizer), training NLL, and validation NLL versus epoch; best validation epoch is annotated. (b) Posterior density with training-data overlay: Heat map of $p(t|x)$ across the $x$–$t$ plane with a thinned set of $(x, t)$ pairs superimposed; multiple ridges at fixed $x$ recover the many-to-one inverse geometry and show that mass is not spuriously "filled" in low-density valleys. (c) Wavefunction components at a representative $x^\star$: Real and imaginary parts over $t \in [-2,2]$ at a representative input $x = -0.564$ from the interior of the observed range; relative phase structure foreshadows the modal pattern of $|\psi|$. (d) Posterior samples + PDF at an $x^\dagger$ chosen by maximal posterior entropy: histogram of 1,200 samples drawn from $p(t|x^\dagger)$ overlaid with the learned PDF; vertical lines mark posterior mean and median. (e) Coefficient magnitudes $|c_k(x^\star)|$ across spectral orders ($k = 0, \ldots, 24$); decay with $k$ indicates spectral smoothness while retained higher orders encode sharper/multi-modal features. (f) Coefficient phases $\arg c_k(x^\star)$ across the same orders. Phase organization across orders governs constructive/destructive interference among $\varphi_k$, setting the number, locations, and sharpness of posterior modes for the same $x$.

Although the overall (global) phase of $\psi$ is unidentifiable, the relative phase between spectral components governs interference and thereby shapes $|\psi|^2$. Smooth variation of $\text{Re}(\psi)$ and $\text{Im}(\psi)$ across $t$ signals an adequately regularized expansion and foreshadows well-localized posterior modes. Figures 2(e) and 2(f) quantify the spectral representation at the same $x^\star$. The decay of $|c_k(x^\star)|$ with order $k$ reflects the effective smoothness of $\psi$ and provides an interpretable complexity measure: sharper or more closely spaced modes generally require slower decay and some high-order energy. The phase profile $\arg c_k(x^\star)$ encodes how basis functions combine constructively or destructively; structured phase progression is characteristic of cleanly separated modes, while erratic phase behavior would suggest overfitting or unnecessary oscillation. Together, magnitudes and phases reveal how the model achieves multi-modality: not by enumerating mixture components, but by spectral superposition whose squared modulus is automatically non-negative and normalizable. Figure 2(d) evaluates local calibration at $x^*$, selected by maximal posterior entropy to stress-test ambiguity. The histogram of 1,200 samples—drawn from the discrete, trapezoid-weighted CDF of the learned PDF—aligns with the analytic curve, supporting both (i) sampling fidelity of the numerical procedure and (ii) accurate allocation of modal mass. The posterior mean and median, marked by vertical lines, typically fall between modes in such settings; this underscores the importance of reporting full densities for decision-making, rather than relying on single-point predictors that may lie in low-density regions.

To provide an insight into the representative power of density approximations based on SNNs, figure 3 (panels a–d)) juxtaposes four non-monotone forward maps (top row) with their corresponding SNN posterior reconstructions $p(t|x)$ (bottom row), revealing that the spectral, analytically normalized wavefunction representation captures both the count and the allocation of modes across $x$. In P1 (winged-sine), figure 3(a), the SNN resolves the fine, ripple-induced alternation between two- and three-modal regions; in P2 (tri-fold), figure 3(b), the SNN reproduces the expected three dominant branches that merge toward the tails; in P3 (mixed harmonics), figure 3(c), the SNN preserves asymmetric mass across competing preimages; and in P4 (modulated five-fold), figure 3(d), the SNN tracks the envelope-driven drift and occasional coalescence of ridges. Because $|\psi_x|^2$ is normalized exactly via $\|\mathbf{c}(x)\|_2^2$ in a Chebyshev-orthonormal basis, the heat maps are true posteriors (with respect to the Chebyshev measure), enabling a single color scale to compare concentration across problems without per-panel rescaling. The ridge brightness follows the geometry of the forward map and ridge width reflects the noise level, while the finite spectral order ($K = 35$) balances resolution with stability, avoiding spurious splitting. Collectively, figure 3 provides compelling evidence that SNNs produce calibrated, high-fidelity, and mixture-free reconstructions of smoothly varying multimodal densities. Crucially, they preserve both the posterior topology (e.g., mode birth and annihilation at folds) and the quantitative mass distribution, both of which are essential prerequisites for rigorous uncertainty quantification.

The normalization constant $\mathfrak{N} = \int_a^b |\psi(t|x)|^2\,dt$ is essential, as the predicted PDF is given by $p(t|x) = |\psi(t|x)|^2/\mathfrak{N}$. We explored two primary approaches for determining $\mathfrak{N}$ in figure 4, for the inverse problem (21). The first approach, trapezoidal integration, numerically computes the integral on a dense, pre-defined grid. This method offers the advantage of being general and model-agnostic, simplifying integration into the computational graph regardless of basis or domain changes; however, it introduces minor discretization error and integration noise into $\mathfrak{N}$'s calculation, potentially yielding a less smooth gradient as backpropagation relies on fixed grid points. Conversely, utilizing the inherent orthogonality and normalization of the Chebyshev basis allows for exact analytic normalization, where the integral simplifies to a direct sum of the squares of the coefficients: $\mathfrak{N} = \sum_{k=0}^{K} |c_k(x)|^2$. This analytic approach entirely eliminates numerical integration error and noise, resulting in a demonstrably precise and smoother gradient for optimization. The stability confirmed by the training curves (figure 4(a)) verifies that the trapezoidal method provides a sufficiently accurate and stable estimate of $\mathfrak{N}$ for successful SNN training; nevertheless, the substitution of numerical integration with the exact analytic normalization further improves the smoothness of the loss function, potentially accelerating asymptotic convergence and achieving a marginally lower minimal validation NLL, as presented in figure 4(b).

## 6. Necessity of Complex coefficients, gauge invariance, and interference

Allowing the spectral coefficients to be complex, $c_k = a_k + ib_k$, increases the effective degrees of freedom available to shape $p(y|\mathbf{x}) = |\psi|^2$ without increasing the basis size $K$. Formally, $\psi = \sum_{k=0}^{K} c_k \varphi_k$ with $\mathbf{c} \in \mathbb{C}^{K+1}$ spans a complex-linear subspace whose real and imaginary parts together yield a $2(K + 1)$-dimensional real space. After imposing normalization $\int |\psi|^2 d\mu = 1$ (one constraint) and quotienting the global phase (one gauge), the complex model has $2K$ real degrees of freedom, whereas the real-only model has $K$ (one constraint for normalization). Thus, at fixed $K$, complex coefficients inject an additional $K$ phase degrees of freedom —precisely the $K$ independent relative phases—enhancing expressivity without enlarging the basis. While the observable density remains a real, nonnegative scalar. Two structural consequences flow from these simple facts and determine both representation power and the design of training procedures. First, because probabilities depend only on the squared modulus of the amplitude, a global phase transformation $e^{i\phi}\psi$ is unobservable in the predictive distribution; accordingly, any loss function, regularizer, or diagnostic must be gauge-invariant with respect to this U(1) action, and optimization dynamics that ignore the gauge freedom risk ill-conditioning or ambiguous parameter trajectories unless they are rendered phase-insensitive by construction (e.g., by working with normalized coefficient vectors on the complex projective sphere or by imposing invariants in the objective). Second, relative phases among the coefficients materially affect cross terms in $|\psi|^2$ and therefore provide a powerful mechanism for spatial reallocation of mass: when the synthesis basis $\{\varphi_k\}$ is real-valued (as with Chebyshev polynomials on a real interval), we have the algebraic expansion

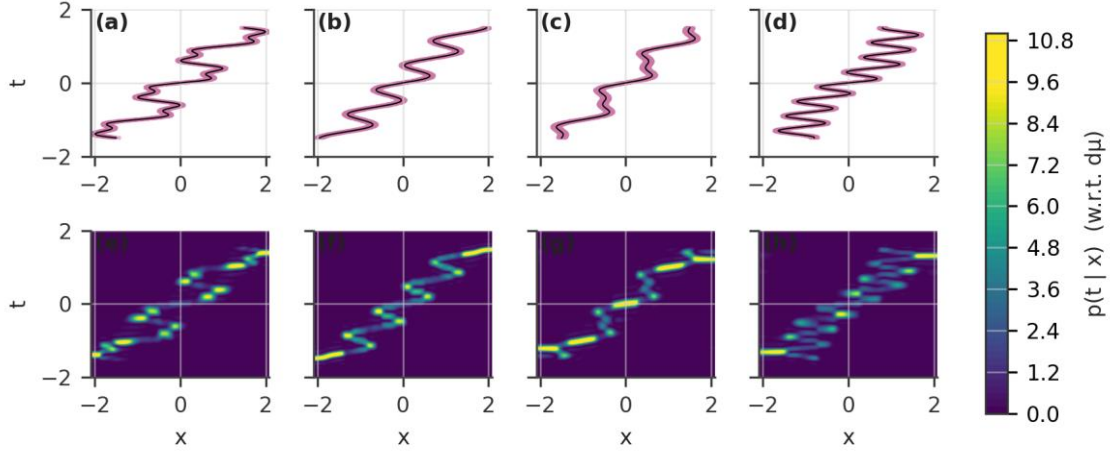

**Figure 3.** Two-row comparison of multimodal inverse problems (top) and SNNs posterior reconstructions (bottom). Panels (a–d): forward relations between the latent variable $t \in [-2,2]$ and the measured response $x$ for four benchmark problems, shown as noisy scatter (uniform noise $\epsilon \sim U[-0.1, 0.1]$) together with the deterministic map (black line, $\epsilon = 0$). Problems are: P1 winged-sine $x = t + 0.60 \sin(2\pi t) + 0.25 \sin(6\pi t)$; P2 tri-fold $x = t + 0.45 \sin(3\pi t)$; P3 mixed harmonics $x = t + 0.35 \sin(2\pi t) + 0.15 \sin(4\pi t)$; P4 modulated five-fold $x = t + 0.50 \sin(5\pi t) + 0.20 \sin(\pi t)$. Bottom panels: SNN estimates of the conditional density $p(t|x)$ learned from 10,000 $(x, t)$ pairs for each problem, rendered as heat maps on a shared $x$–$t$ grid. Densities are with respect to the Chebyshev measure. A single color scale is shared across bottom panels to enable direct quantitative comparison of modal structure and concentration.

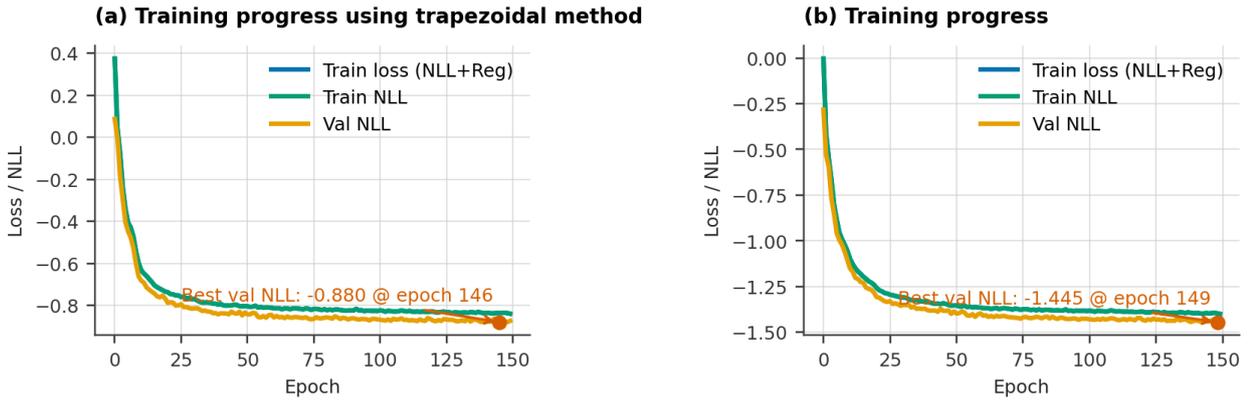

**Figure 4.** Comparison of normalization strategies for the SNN with the non-invertible inverse map (21). (a) Training behavior using numerical trapezoidal integration to compute the normalization constant $\mathfrak{N} = \int_a^b |\psi(t|x)|^2 \, dt$. (b) Training using the analytic Chebyshev normalization $\mathfrak{N} = \sum_{k=0}^K |c_k(x)|^2$: exploiting basis orthogonality removes numerical integration error and gradient noise, yielding a noticeably smoother loss trajectory and a marginally lower asymptotic validation negative log-likelihood, consistent with faster asymptotic convergence. Panels together demonstrate the tradeoff between generality (numerical trapezoid) and precision/smoothness (analytic Chebyshev), and quantify the practical benefit of analytic normalization when an orthonormal spectral basis is available.

$$|\psi|^2 = \sum_k |c_k|^2 \varphi_k(y)^2 + 2 \sum_{j<k} \left( \text{Re}(c_j)\text{Re}(c_k) + \text{Im}(c_j)\text{Im}(c_k) \right) \varphi_j(y) \varphi_k(y)$$
$$= \sum_k |c_k|^2 \varphi_k(y)^2 + 2 \sum_{j<k} |c_j||c_k| \cos(\theta_j - \theta_k) \varphi_j(y) \varphi_k(y), \quad (22)$$

where $c_k = |c_k| e^{i\theta_k}$. The second sum is an explicit interference term. This interference is not merely algebraic ornamentation but the central representational lever by which an SNN sculpts multimodal, skewed, or heavy-tailed conditional shapes using a compact spectral budget. With coefficient magnitudes $|c_k|$ held fixed, varying the phases $c_k = |c_k| e^{i\theta_k}$ reshapes the density via the cross terms $2 \sum_{j<k} |c_j||c_k| \cos(\theta_j - \theta_k) \varphi_j(y) \varphi_k(y)$. Because different basis pairs $\varphi_j(y) \varphi_k(y)$ have distinct nodal and oscillatory patterns, choosing relative phases $\theta_j - \theta_k$ to make $\cos(\theta_j - \theta_k)$ positive (negative) amplifies (attenuates) density exactly where $\varphi_j \varphi_k$ is positive, and the reverse where it is negative. For an orthonormal basis the total mass, $\int |\psi|^2 d\mu = \sum_{k=0}^K |c_k|^2$, is phase-invariant, holding the amplitudes $|c_k|$ fixed keeps the total probability mass fixed no matter how you change $\{\theta_k\}$, so any local increase must be offset elsewhere. This "phase-only" control provides a parsimonious mechanism to form narrow peaks, deep valleys, or asymmetric lobes—within the resolution permitted by the truncation $K$—without adding new basis components or increasing $K$. If $c_k \in \mathbb{R}$, the cross terms in $|\psi|^2$ are $2 \sum_{j<k} c_j c_k \varphi_j(y) \varphi_k(y)$. Writing $c_k = \sigma_k |c_k|$ with

$\sigma_k \in \{\pm 1\}$, the pairwise factor becomes $\sigma_k \sigma_j |c_k| |c_j|$. In phase language this is $|c_k| |c_j| \cos \Delta \theta_{jk}$ with $\theta_{jk} \in \{0, \pi\}$ and $\cos \Delta \theta_{jk} \in \{\pm 1\}$. This restricts the interference factor to $\pm 1$. Hence, with real-only coefficients, to change a cross term you typically must also change $|c_k|$ or $|c_j|$, which automatically perturbs the diagonal terms $|c_k|^2 \varphi_k^2$ or $|c_j|^2 \varphi_j^2$. By contrast, complex coefficients $c_k = |c_k| e^{i\theta_k}$ allow $\cos \Delta \theta_{jk} \in [-1,1]$ to vary continuously while holding $|c_k|$ fixed, so cross-term interference can be tuned independently of the diagonal contributions $|c_k|^2 \varphi_k(y)^2$. The loss of this phase-only control in a real-only architecture makes fine redistribution of probability at fixed spectral order more difficult: to achieve comparable shape diversity one typically needs (i) a larger $K$ so more basis products can combine through amplitudes alone, or (ii) auxiliary mechanisms (e.g., explicit mixture components). Both routes increase parameter counts and usually complicate optimization.

In practice, therefore, the choice to use complex coefficients should be guided by the interplay between expected conditional complexity and desiderata for calibration and interpretability. For one- to few-dimensional outputs where multimodality, asymmetric peaks, or delicate valley structures are common, complex coefficients provide an efficient mechanism for capturing such features with a controlled parameter budget; for very high-dimensional outputs or cases with extreme non-smoothness, the benefits are attenuated and alternative strategies (separable factorization, localized bases) may be preferable.

Under identical training data, basis truncation, and quadrature, the consequences of allowing complex amplitudes in the CSNN are most clearly visible in the learned posterior geometry: the CSNN more faithfully recovers the multi-branch structure implied by the forward map and, in heat-map visualizations (figure 5(a), for the inverse problem (21)), traces high-density ridges that follow the true branches and reproduce the correct modal count. Mechanistically, complex coefficients introduce spectral phase degrees of freedom that enable constructive and destructive interference among basis functions; this (i) preserves separation between nearby branches without inflating density on intervening regions, (ii) reduces mode-displacement errors by steering peaks via sub-grid phase adjustments rather than coarse reweighting, and (iii) allocates probability mass across competing lobes in a manner that respects subtle shape cues. Because the conditional PDF is the squared modulus of the complex wavefunction, these phase adjustments let the network form sharply peaked, distinct modes without forcing coefficient magnitudes alone to encode all structure, yielding more accurate basin-wise mass compared with the ground truth. By contrast, a RSNN — trained on the same data and basis — must typically increase the number of basis terms to approximate multimodality; this tends to be less efficient and often produces ripples, spurious shoulders, overly rigid peak shapes, or weak off-support density as the model compensates for the absence of relative phase (figure 5(b)), thereby degrading resolution in transition regions where the forward map is non-monotone or heteroscedastic.

While the heat map offers a synoptic view of the two-dimensional posterior, figure 6—a grid of slice plots for the inverse problem (21)—provides the substrate for statistically precise follow-ups and enables detailed inspection of local modal geometry. Read horizontally and vertically, the slices first illuminate local modal structure: by comparing peak locations, heights, and widths, one can assess whether the complex-coefficient and real-coefficient heads allocate mass to the same modes with comparable sharpness. Instances in which the complex head produces higher, narrower peaks while the real head distributes probability more broadly indicate that phase interactions in the complex expansion effect constructive interference at modal centers and destructive interference in surrounding regions; conversely, near-coincident curves imply that additional phase degrees of freedom do not materially improve the local fit at that particular **x**. The slices further expose subtle inter-modal features that are easily obscured in a heat-map integration—small shoulders, asymmetric tails, or incipient multimodality often appear as slight but systematic separations between the two curves. These can be quantified per slice via Kullback–Leibler divergence (KL distance) [37-39], squared ($\ell_2$) distance [40], or differences in low-order moments (mean, variance, skewness) [41] to establish statistical significance. Finally, progressing left-to-right and top-to-bottom in the grid reveals how local discrepancies evolve with **x**. Frequently, pronounced differences concentrate in transition regions where the forward map is non-monotone or multi-valued and are negligible elsewhere, reinforcing the interpretation that the models diverge chiefly where expressive demand is highest.

## 7. Operator Postulate: Observables and Measurement in the SNN

The SNN already derives its probabilistic semantics from two pillars of quantum theory—de Broglie and Born principles. A natural next step is to extend this foundation with a third postulate patterned on the operator–observable structure of quantum mechanics: we posit that to every application-relevant measurable quantity on the output space there corresponds a self-adjoint operator acting on the amplitude Hilbert space, and that empirical queries, constraints, and decision criteria can be expressed as expectation values, variances, and spectral properties of these operators with respect to $\psi_\mathbf{x}$. Formally, let $\mathcal{H} = L^2(\mathcal{Y}, \nu)$ be the Hilbert space of square-integrable complex functions on the output domain $\mathcal{Y}$ with respect to a base measure $\nu$ (counting measure for discrete labels, Lebesgue for continuous), and let $\psi_\mathbf{x} \in \mathcal{H}$ be normalized, $\langle \psi_\mathbf{x}, \psi_\mathbf{x} \rangle$. We postulate: (i) every observable $A$ that we may care to interrogate about the output—location, dispersion, threshold exceedance, regime indicators, or domain-specific functionals—corresponds to a densely defined self-adjoint operator $\hat{A}$ on $\mathcal{H}$; (ii) the predicted value of $A$ at input **x** is the expectation

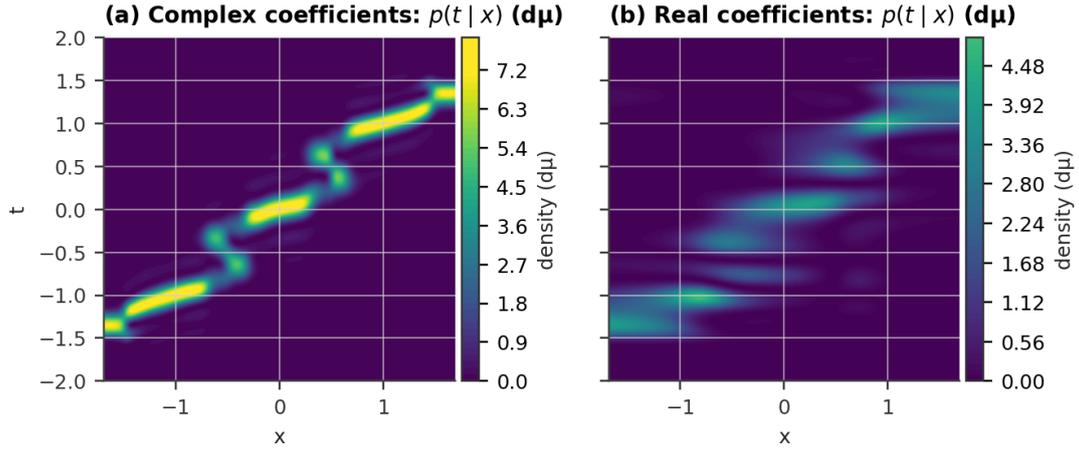

**Figure 5.** Comparison of conditional PDFs estimated by SNNs with complex vs. real Chebyshev coefficients (side-by-side conditional density heatmaps) for the non-invertible inverse map (21). (a) estimated conditional density $p(t \mid x)$ produced by the SNN whose last layer emits complex coefficients; (b) the analogous heatmap for the SNN variant constrained to real coefficients. Both panels show the same plotting domain, with the horizontal axis giving the observed covariate $x$ (the input grid used for evaluation) and the vertical axis giving the latent/target variable $t$ (the Chebyshev-mapped output grid).

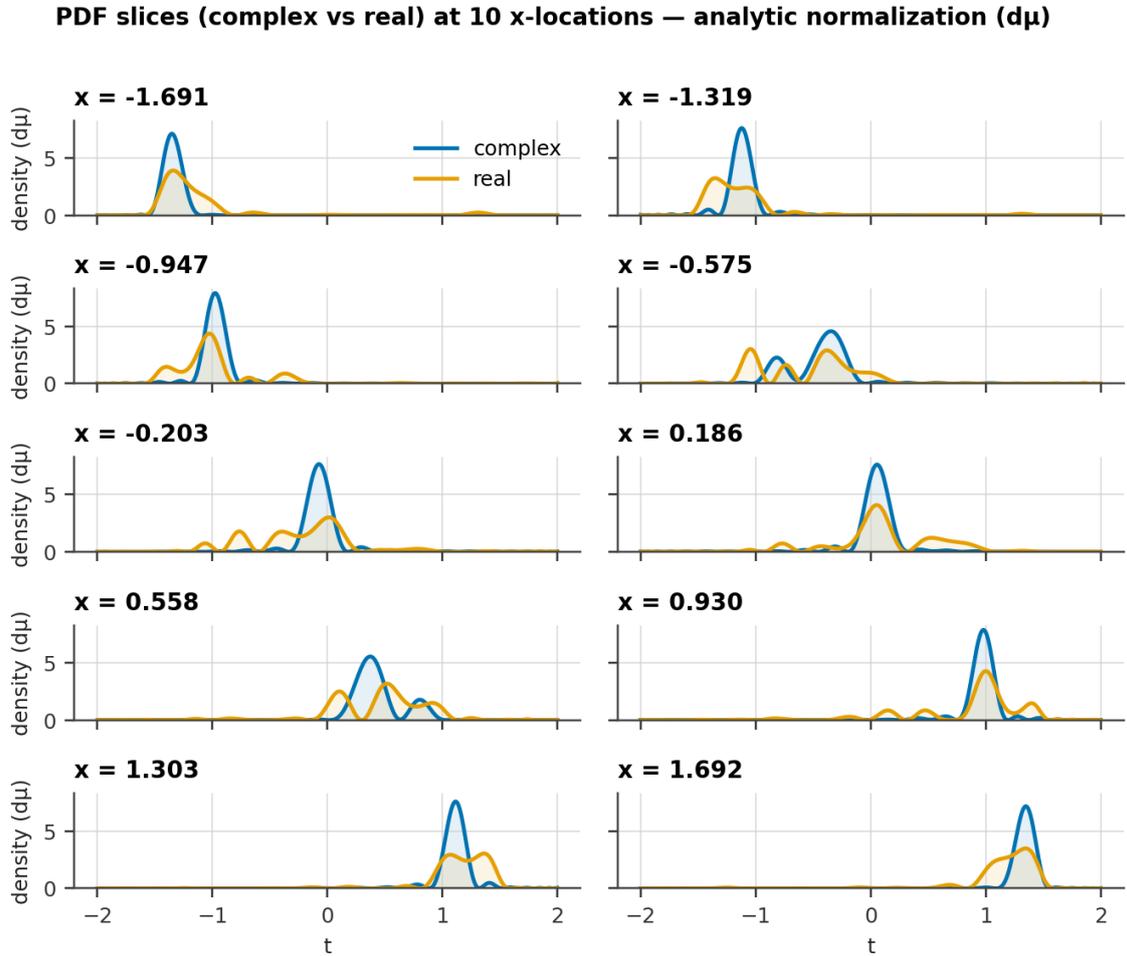

**Figure 6.** PDF slices (complex vs real) at selected $x$-locations for the non-invertible inverse map (21). The figure displays ten vertically oriented subplots; each panel overlays the one-dimensional conditional probability density $p(t \mid x_j)$ for the CSNN (blue) and the RSNN (yellow) at a specific $x_j$ chosen evenly across the evaluated $x$-grid. The horizontal axis in each subplot is the target variable $t$ and the vertical axis is the density value; the plotting routine enforces a common $y$-axis limit across panels so that direct visual comparisons of peak height and tail mass are fair. Each subplot title displays the numeric value of $x_j$ so that slices can be mapped back to features in the heatmaps.

$$\mathbb{E}[A|\mathbf{x}] = \langle \psi_{\mathbf{x}}, \hat{A}\psi_{\mathbf{x}} \rangle, \tag{23}$$

while the associated predictive dispersion is

$$\mathrm{Var}[A|\mathbf{x}] = \langle \psi_{\mathbf{x}}, (\hat{A} - \mathbb{E}[A])^2 \psi_{\mathbf{x}} \rangle, \tag{24}$$

and (iii) when two observables are represented by non-commuting operators, their simultaneous predictability is constrained by uncertainty relations inherited from the Robertson–Schrödinger inequality. This operator postulate does not alter the Born semantics; rather, it supplies a rigorous calculus that elevates the SNN from "a model of densities" to "a model of observables with a consistent algebra," providing a unifying language for training objectives, regularization, symmetry, calibration, and decision-making that is mathematically coherent and computationally tractable in the spectral parameterization already used to construct $\psi_{\mathbf{x}}$.

The simplest and most pervasive class of observables consists of multiplication operators. For any measurable function $o: \mathcal{Y} \to \mathbb{R}$, define

$$(\hat{o}\psi)(y) = o(y)\psi(y). \tag{25}$$

For such operators, the spectral theorem is trivial—the spectrum is the essential range of $o$—and expectation values reduce to classical integrals:

$$\mathbb{E}[o(y)|\mathbf{x}] = \int o(y)\,|\psi_{\mathbf{x}}|^2 d\nu(y) = \langle \psi_{\mathbf{x}}, \hat{o}\psi_{\mathbf{x}} \rangle. \tag{26}$$

Thus, any statistical functional that can be written as an expectation of a function $o(y)$ is immediately represented by $\hat{o}$. So the quantum-style inner product with $\hat{o}$ is the classical expectation of $o(y)$ with respect to the density $p(y|\mathbf{x}) = |\psi_{\mathbf{x}}(y)|^2$.

For example, the mean (conditional expectation) is, by definition, the first moment of the distribution — i.e., the average of the outcome weighted by its probability density. Concretely, if the predictive density for $y|\mathbf{x}$ is $p(y|\mathbf{x})$, the conditional mean is

$$\mathbb{E}[y|\mathbf{x}] = \int y\, p(y|\mathbf{x}) d\nu(y). \tag{27}$$

In the SNN, the density is given by the Born map $p(y|\mathbf{x}) = |\psi_{\mathbf{x}}(y)|^2$ (with respect to base measure $\nu$), so substituting this identity yields

$$\mathbb{E}[y|\mathbf{x}] = \int y\,|\psi_{\mathbf{x}}(y)|^2 d\nu(y). \tag{28}$$

That integral is the inner product (in $L^2(\mathcal{Y}, \nu)$) of $\psi_{\mathbf{x}}$ with the multiplication operator $\hat{y}$ applied to $\psi_{\mathbf{x}}$:

$$\langle \psi_{\mathbf{x}}, \hat{y}\psi_{\mathbf{x}} \rangle. \tag{29}$$

Thus, taking $o(y) = y$ corresponds exactly to the mean because the expectation of the function $y$ under the predictive distribution is, by definition, the mean. Another example is the second moment and variance. Take $o(y) = y^2$. The second moment is

$$\mathbb{E}[y^2|\mathbf{x}] = \int y^2\,|\psi_{\mathbf{x}}(y)|^2 d\nu(y) = \langle \psi_{\mathbf{x}}, \widehat{y^2}\psi_{\mathbf{x}} \rangle. \tag{30}$$

The variance is then the difference of operator expectations:

$$\mathrm{Var}[y|\mathbf{x}] = \int y^2\,|\psi_{\mathbf{x}}(y)|^2 d\nu(y) - \left(\int y\,|\psi_{\mathbf{x}}(y)|^2 d\nu(y)\right)^2 = \langle \psi_{\mathbf{x}}, \widehat{y^2}\psi_{\mathbf{x}} \rangle - \langle \psi_{\mathbf{x}}, \hat{y}\psi_{\mathbf{x}} \rangle^2. \tag{31}$$

Equivalently,

$$\mathrm{Var}[y|\mathbf{x}] = \langle \psi_{\mathbf{x}}, (\hat{y} - \mathbb{E}[y|\mathbf{x}])^2 \psi_{\mathbf{x}} \rangle. \tag{32}$$

Threshold exceedance can be represented as an indicator observable. For a prescribed threshold $t \in \mathcal{Y}$, define the multiplication operator $(\hat{\mathbf{1}}_{(t,\infty)}\psi)(y) = \mathbf{1}\{y > t\}\psi(y)$, where $\mathbf{1}\{\cdot\}$ denotes the indicator function. The corresponding operator expectation under the normalized amplitude $\psi_{\mathbf{x}}$ equals the conditional exceedance probability:

$$\mathbb{E}[\mathbf{1}\{y > t\}|\mathbf{x}] = \int \mathbf{1}\{y > t\}\,|\psi_{\mathbf{x}}(y)|^2 d\nu(y) = \Pr(y > t|\mathbf{x}) = \langle \psi_{\mathbf{x}}, \hat{\mathbf{1}}_{(t,\infty)}\psi_{\mathbf{x}} \rangle. \tag{33}$$

Thus, probabilities of events are expectations of indicator-operators, providing a direct bridge between event-level queries and the operator calculus.

Moreover, if $\{\varphi_k\}$ (e.g., Chebyshev atoms on a compactified coordinate) is the orthonormal basis used to represent $\psi_{\mathbf{x}}(y)$, then the expectation of any real-valued observable $o$ (multiplication operator) is a Hermitian quadratic form in the coefficient vector $\mathbf{c}(\mathbf{x})$:

$$\langle \psi_{\mathbf{x}}, \hat{o}\psi_{\mathbf{x}} \rangle = \sum_{j,k} c_j^*(\mathbf{x}) F_{jk} c_k(\mathbf{x}) = \mathbf{c}^*(\mathbf{x}) F \mathbf{c}(\mathbf{x}), \qquad F_{jk} = \langle \varphi_j, \hat{o}\varphi_k \rangle = \int \varphi_j(y) o(y)\, \varphi_k(y) d\nu(y). \tag{34}$$

This identity is important for practice: once the matrix $F$ for a chosen observable is computed (by analytic integration or numerical quadrature), evaluating the expectation and its gradients with respect to the coefficients is cheap and differentiable. In this sense, the operator postulate subsumes and extends the SNN's probabilistic semantics because any statistic that is a measurable functional of $y$ is an expectation of a multiplication operator. It also provides a disciplined route to multi-objective training: instead of optimizing just the NLL, $-\sum_i \log|\psi_{\mathbf{x}_i}(y_i)|^2$, one may add penalty terms that are expectations of operators reflecting domain constraints or desiderata—tail-control via $\hat{o}$ for large $|y|$, safety margins via $\hat{o}$ that spikes beyond limits, or calibration shapers via entropy-like functionals recast as quadratic forms through $\psi$. In general

$$\mathcal{L} = \mathcal{L}_{NLL}(\theta) + \sum_j \lambda_j \frac{1}{N} \sum_{i=1}^{N} R_j(\psi_{\mathbf{x}_i}), \qquad R_j(\psi_{\mathbf{x}}) = \langle \psi_{\mathbf{x}}, \hat{o}_j \psi_{\mathbf{x}} \rangle, \tag{35}$$

where $\lambda_j \geq 0$ are penalty weights. Each $\hat{o}_j$ encodes a desideratum (tail mass, safety, calibration, etc.).

## 8. Kinetic and potential quadratic regularizers and their uncertainty interpretation

The SNN represents conditional densities through a complex-valued wavefunction expanded in a functional basis such as Chebyshev polynomials or Fourier modes. In this representation, each basis element carries a characteristic oscillation scale: low-index elements produce broad, slowly varying shapes while high-index elements oscillate rapidly and supply the short wavelengths needed to resolve fine features. Consequently, a conditional distribution with a very narrow peak or a sharp edge can only be synthesized by combining sufficiently many high-order terms — in other words, by increasing the model's spectral complexity, which quantifies how much of the function's energy lives at high frequencies. Allowing high spectral complexity grants expressive power and the ability to form near-deterministic, sharply localized predictions, yet it also raises the risk of overfitting because the same high-frequency capacity can reproduce noise as narrow spikes. Conversely, low spectral complexity tends to improve stability and out-of-sample calibration but cannot capture abrupt structure. Framing this problem in operator terms explains this trade-off clearly: increased localization necessarily entails increased spectral content, so choices about basis truncation and spectral regularization amount to deliberate decisions about the acceptable balance between sharpness and smoothness in the learned conditional densities.

Beyond multiplication, differential and shift operators introduce structure not captured by pointwise functionals, and their commutators anchor uncertainty principles that translate directly into regularizers and diagnostics. On a continuous scalar output, define the "position" operator $\hat{Y}$ by $(\hat{Y}\psi)(y) = y\psi(y)$ and the "frequency" or "momentum" operator $\hat{P} = -i\partial_y$ on a suitable domain (with boundary conditions induced by the compactification used to map $\mathbb{R}$ to $[-1,1]$). These satisfy

$$[\hat{Y}, \hat{P}] = i\hat{I}, \tag{36}$$

up to boundary terms. These operators are not mere abstract constructs; when applied to the SNN wavefunction, they capture two complementary aspects of predictive behavior. The position operator encodes where the wavefunction concentrates mass in the output domain and therefore directly relates to the sharpness and localization of the predictive density. The momentum operator encodes how rapidly the wavefunction oscillates across the output domain and therefore relates to the spectral content (spectral complexity) of the density.

Crucially, localization and spectral content are coupled by the Robertson–Schrödinger inequality, which for these operators reads

$$\mathrm{Var}(\hat{Y})\mathrm{Var}(\hat{P}) \geq \frac{1}{4}|\langle[\hat{Y}, \hat{P}]\rangle|^2 + \frac{1}{4}|\langle\{\Delta\hat{Y}, \Delta\hat{P}\}\rangle|^2 = \frac{1}{4} + [\mathrm{Cov}_{\mathrm{sym}}(\hat{Y}, \hat{P})]^2, \tag{37}$$

where $\Delta\hat{A} = \hat{A} - \langle\hat{A}\rangle$ and $\mathrm{Cov}_{\mathrm{sym}}$ is the symmetrized covariance. When the covariance between position and momentum vanishes or is negligible, the bound reduces to the familiar Heisenberg-like limit that the product of the position and momentum uncertainties cannot be smaller than a fixed minimal value;

$$\Delta Y_{\mathbf{x}} \Delta P_{\mathbf{x}} \geq \frac{1}{2}, \tag{38}$$

when the covariance vanishes; here $\Delta A_{\mathbf{x}}^2 = \mathrm{Var}[A|\mathbf{x}]$. Practically, this means that a wavefunction that is extremely sharply localized in the output domain necessarily contains substantial high-frequency content; conversely, a wavefunction with strictly low-frequency content must be relatively delocalized in the output coordinate. The bound thus encodes a precise trade-off between sharpness and spectral complexity that any SNN must respect by virtue of basic functional analysis, independent of training details. In other words, the uncertainty bound explains why there is no free lunch: extreme output localization requires spectral complexity, and limiting spectral complexity through regularization will necessarily limit the sharpness of conditional densities the model can represent.

A second practical consequence relates to the choice of basis and truncation. SNNs commonly use Chebyshev polynomials or other orthogonal bases on the compactified domain because of their spectral convergence properties for smooth targets. Spectral bases have favorable approximation rates for functions that are sufficiently smooth: the error in approximating a smooth function decays rapidly with the number of basis elements. However, that spectral accuracy presumes the function does not require excessive high-frequency energy to represent localized features. When the target conditional density possesses narrow peaks, sharp edges, or near-singular features, the approximation demands many high-order basis functions, inflating momentum uncertainty. Truncating the basis to a finite number reduces computational cost and regularizes the problem, but also implements an implicit low-pass filter that enforces spectral simplicity and therefore spreads out any features in the output domain. From the uncertainty viewpoint, this spreading is not an artifact but a necessary corollary of suppressing high-frequency components: one may reduce momentum uncertainty only by allowing position uncertainty to increase.

Figures 7 and 8 systematically investigate the influence of the Chebyshev truncation order $K$ on the expressive capability and stability of the CSNN in modeling conditional densities $p(t|x)$ for the inverse problem (21). Because all models are trained on the same dataset and evaluated over an identical $x$-grid, differences between panels can be attributed primarily to the spectral representational capacity and its interaction with the chosen regularizer, optimizer warm-up, and early-stopping policy.

Figure 7 presents heat maps of the posterior distributions over the latent variable $t$ for a range of observed inputs $x$, with each panel corresponding to a distinct truncation order $K$ spanning from 10 to 35. The densities are normalized analytically via the exact $\ell_2$-norm of the Chebyshev coefficients, ensuring consistency with the orthonormal spectral basis and preserving the integral under the Chebyshev measure $d\mu$. A uniform color scale is applied across all panels, guaranteeing that variations in apparent sharpness or contrast reflect true redistribution of probability mass rather than plotting artifacts. Interpreting the grid proceeds at both global and local levels. At the global level, one evaluates the stability of major features, including mode locations that follow the forward map, heteroscedastic spread, and the presence of multimodality. Stable representation across $K$ indicates that the network has sufficient capacity to capture the essential structure and that further increasing $K$ yields diminishing returns. At the local level, one inspects finer-scale features such as narrower peaks, additional shoulders or submodes, and small oscillatory fringes, which emerge only when higher-frequency basis functions are included. The panel design makes it straightforward to detect the classical underfitting/overfitting trade-off encountered in spectral methods: too small $K$ will smooth over multimodality and cannot position sharp peaks accurately, whereas excessive $K$ can permit high-frequency interference and Gibbs-like ringing in regions where data are sparse. Inspection of the heat maps reveals clear trends. For low truncation orders ($K = 10$ and $K = 15$), posterior densities are relatively smooth and under-resolve secondary peaks in locally non-invertible regions, producing broadened distributions. As $K$ increases, the CSNN progressively captures sharper peaks and emergent multimodality, with higher-order modes contributing primarily to subtle refinements rather than qualitative changes beyond $K = 25$. This behavior is consistent with the spectral accuracy of Chebyshev expansions for smooth densities, where truncation error decreases rapidly with $K$. Figure 8 complements these heat maps by displaying conditional density slices at ten representative $x$-locations, overlaid across all considered $K$ values. This view clearly illustrates how modal structure evolves with increasing $K$. Low-order truncations produce diffuse and partially merged modes, whereas higher $K$ resolves secondary peaks, maintaining their relative amplitude and separation.

Quadratic regularizers make the trade-off controllable during training by penalizing functionals of $\psi_\mathbf{x}$ that correlate with spectral content and spatial preference. We introduce two complementary regularizers. The first is a kinetic-energy penalty, defined for each input $\mathbf{x}$ as

$$\mathcal{E}_{\text{kin}}(\mathbf{x}) = \lambda_{\text{kin}} \int_a^b |\partial_y \psi_\mathbf{x}(y)|^2 \, dy = \mathbf{c}^*(\mathbf{x})\mathbf{K}\mathbf{c}(\mathbf{x}), \tag{39}$$

is the Dirichlet form associated with the Laplacian, with stiffness matrix $K_{jk} = \int_a^b \varphi_j'(y)\varphi_k'(y) \, dy$. $\mathcal{E}_{kin}$ penalizes large gradients and, therefore, high-frequency oscillations and over-sharpened peaks. The uncertainty inequality then predicts the corollary effect on position uncertainty: as one tightens the spectral penalty, the model will favor broader predictive densities. If the downstream objective demands sharp localization, for instance, to represent nearly deterministic responses, then the practitioner must either relax the spectral penalty or increase the spectral truncation budget. Conversely, if calibrated predictive variances and smoothness across inputs are more important than reproducing extremely narrow modes, then a strong spectral penalty is aligned with the modeling goal and produces well-behaved conditional densities.

A complementary regularizer focuses on the potential energy of the wavefunction,

$$\mathcal{E}_{\text{pot}}(\mathbf{x}) = \lambda_{\text{pot}} \int_a^b V(y)|\psi_\mathbf{x}(y)|^2 \, dy = \mathbf{c}^*(\mathbf{x})\mathbf{M}\mathbf{c}(\mathbf{x}), \qquad V(y) \geq 0, \tag{40}$$

with a nonnegative shaping function $V$ and matrix $\mathbf{M}$ has entries $M_{jk} = \int_a^b V(y)\varphi_j(y)\varphi_k(y) \, dy$. Designing potentials that favor concentration in regions of interest can be an efficient way to achieve localization without grossly inflating global spectral energy: a convex well (e.g., $V(y) \propto (y - \mu)^2$) encourages concentration near $\mu$; barrier functions that rise steeply near $a$ or $b$ discourage boundary-hugging tails; multiwell or asymmetric forms can softly encode prior structure (e.g., avoid gaps, favor plausible ranges). In practice, potential design can be learned or engineered: one may use a fixed potential chosen to encode prior beliefs about likely output ranges, or include a learnable potential-like weighting that adapts to data. Either way, the balancing act remains: increasing potential-based localization reduces position uncertainty but must be paid for in spectral energy. Practically, $\mathcal{E}_{\text{pot}}$ complements the kinetic penalty: $\mathcal{E}_{\text{kin}}$ suppresses rapid oscillations (controls smoothness), while $\mathcal{E}_{\text{pot}}$ shapes where mass is allowed to concentrate (controls localization).

Together, potential and kinetic energy regularizers define a Schrödinger-type quadratic operator

$$\mathcal{E}(\mathbf{x}) = \mathbf{c}^*(\mathbf{x})\mathbf{H}\mathbf{c}(\mathbf{x}), \qquad \mathbf{H} = \lambda_{\text{kin}}\mathbf{K} + \lambda_{\text{pot}}\mathbf{M}, \tag{41}$$

which is positive semidefinite and Hermitian. The overall objective combines likelihood and energy,

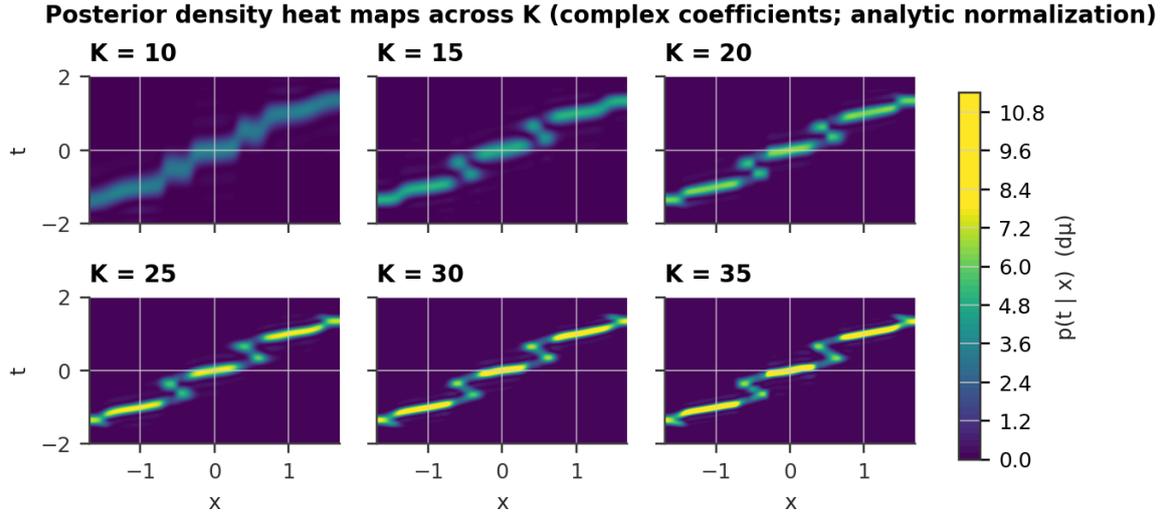

**Figure 7.** Grid of posterior density heat maps across spectral order $K$ for the non-invertible inverse map (21). Posterior densities $p(t \mid x)$ computed by independent CSNNs are shown in a 2x3 panel arranged by increasing Chebyshev expansion order $K \in \{10, 15, 20, 25, 30, 35\}$; each panel plots the same evaluation grid in the observed covariate $x$ (horizontal axis) and the target variable $t$ (vertical axis), with color indicating the per-$x$ conditional density computed on a uniform $t$-grid.

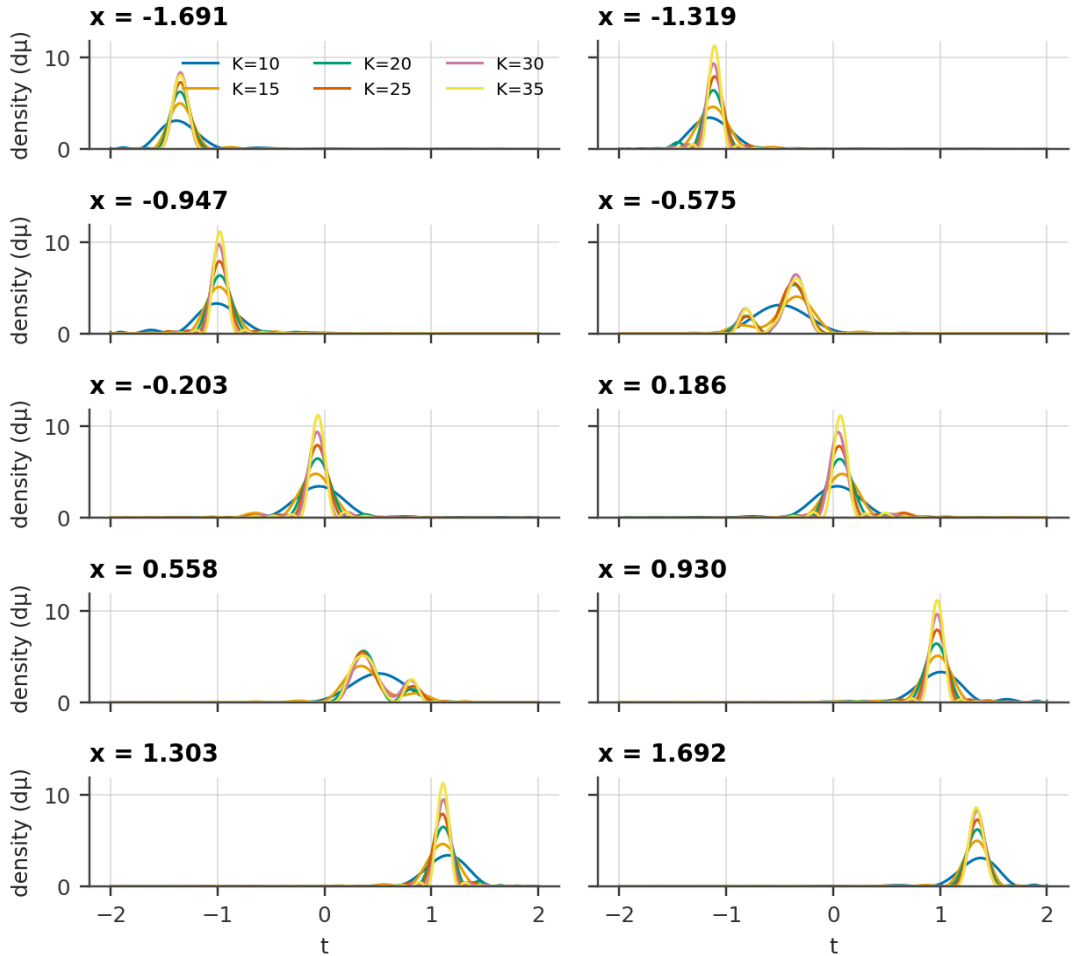

**Figure 8.** Conditional probability density slices at ten representative $x$-locations across the observed domain, for the non-invertible inverse map (21), overlaid for all $K$ values, $K \in \{10, 15, 20, 25, 30, 35\}$, to facilitate comparison of modal structure, amplitude, and resolution.

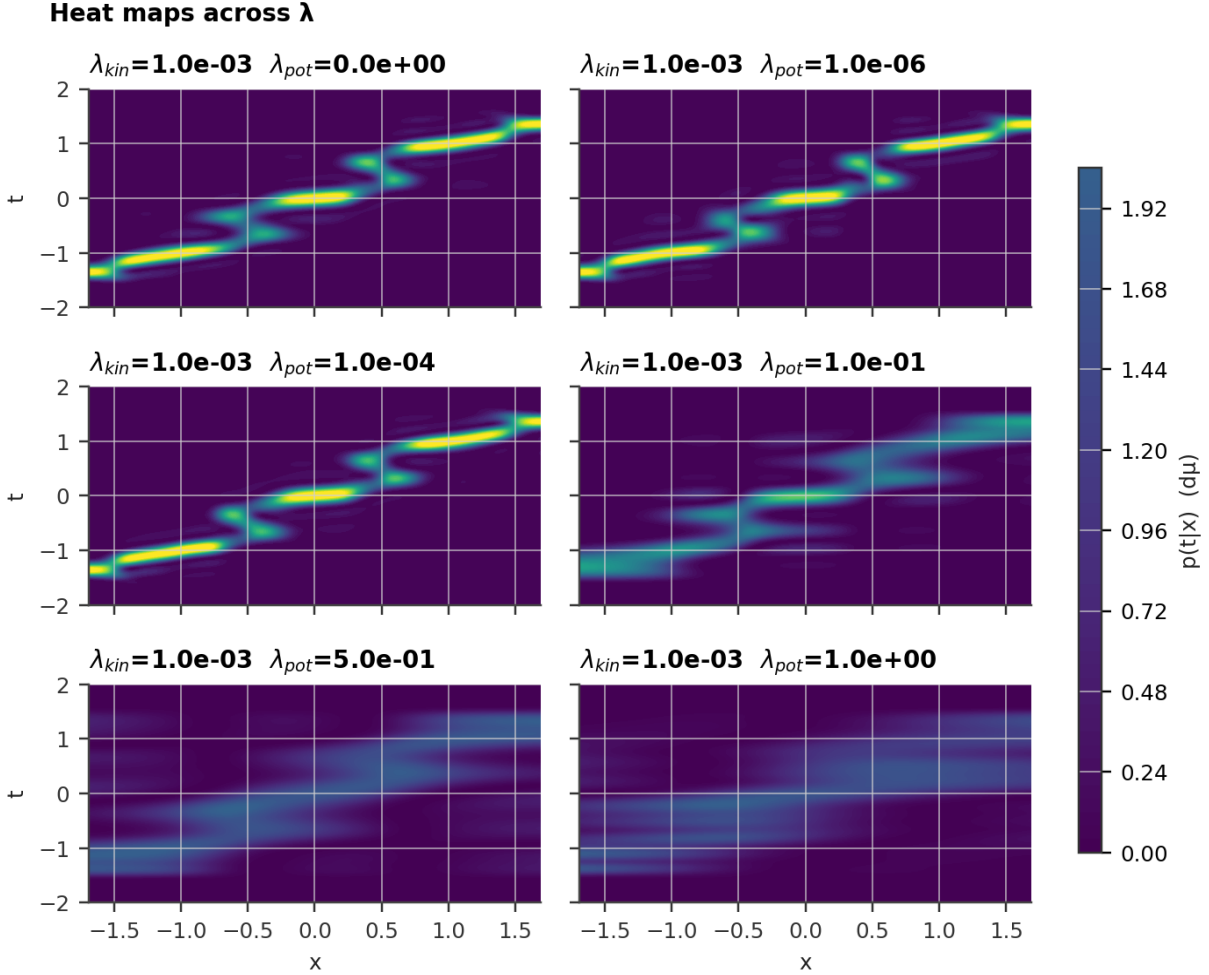

**Figure 9.** Posterior density analysis of a CSNN under varying potential penalty $\lambda_{\text{pot}}$ (using harmonic oscillator potential), for the non-invertible inverse map (21). Heat maps of the conditional probability density $p(t|x)$ across a 3×2 grid of potential penalty strengths $\lambda_{\text{pot}} \in \{0, 10^{-6}, 10^{-4}, 0.1, 0.5, 1\}$, with the kinetic (smoothness) penalty $\lambda_{\text{kin}}$ fixed at $10^{-3}$. Each panel visualizes the conditional density over the latent variable $t \in [-2, 2]$, normalized analytically using the exact Chebyshev coefficient $\ell_2$-norm under the Chebyshev measure $d\mu$.

$$\mathcal{J}(\theta) = \mathcal{L}_{NLL}(\theta) + \mathbb{E}_{\mathbf{x}}[\mathcal{E}(\mathbf{x})], \tag{42}$$

where the expectation over $\mathbf{x}$ is approximated by a minibatch average. Because $\mathcal{E}$ is quadratic in $\mathbf{c}$, it is invariant to global phase, differentiable, and inexpensive to evaluate once $\mathbf{H}$ is precomputed. Tuning $\lambda_{\text{kin}}$ and $\lambda_{\text{pot}}$ or adapting $V$ across tasks, thus offering a principled, physics-inspired knob for controlling both smoothness and localization in SNNs.

One of the most powerful consequences of casting SNN constraints in operator language is clarifying model selection trade-offs. Choose too small a spectral truncation and the model is essentially low-pass; it will produce smooth, reliable densities but will fail to capture narrow modes. Choose too large a truncation without adequate regularization, and the model will overfit, generating spurious oscillations and producing densities with small position variance in training but poor calibration and generalization out of sample. The uncertainty inequality provides a quantitative lens to anticipate these outcomes: increasing the number of basis functions increases the model's ability to independently allocate momentum and position uncertainty, so the same spectral penalty will have different effects depending on truncation. This perspective recommends tuning spectral penalty strength jointly with truncation level, rather than independently, and suggests the use of validation metrics that measure both sharpness and calibration of predictive densities.

Figures 9 and 10 examine the effect of systematically increasing the potential penalty $\lambda_{\text{pot}}$ for the inverse problem (21), on the posterior distributions learned by the CSNN, while maintaining a fixed kinetic (smoothness) penalty $\lambda_{\text{kin}} = 10^{-3}$. The SNN employs analytic normalization of the Chebyshev spectral expansion, ensuring exact probability conservation under the Chebyshev measure $d\mu$. The kinetic penalty provides a baseline level of smoothness, suppressing high-frequency spectral components and limiting oscillatory artifacts, while the potential penalty (harmonic oscillator potential) biases the distribution toward regions of low potential energy. By systematically varying the penalty strengths (potential penalty), the heat maps (figure 9) reveal how the conditional density $p(t|x)$ evolves across the latent-observed domain. Each panel corresponds to a distinct combination of penalty parameters (fixed value of kinetic energy penalty $\lambda_{\text{kin}} = 0.001$ and potential penalty $\lambda_{\text{pot}}$ changes form zero to 1), enabling direct visual assessment of their effects on peak localization, modal structure, and overall smoothness. The use of a shared color scale ensures that apparent differences reflect genuine reallocation of probability mass rather than plotting artifacts. Inspection of the heat maps shows that increasing the potential penalty $\lambda_{\text{pot}}$ biases the distributions towards low-potential regions (in the harmonic oscillator potential, $t = 0$ is low-potential region), concentrating probability mass near the potential minimum while attenuating density in energetically unfavorable regions. Figure 10 provides complementary insights through conditional density slices at ten evenly spaced $x$-locations. Overlaying the densities across all penalty configurations emphasizes how the magnitude and distribution of probability mass shifts with increasing potential penalty, $\lambda_{\text{pot}}$. Importantly, the analytic normalization ensures that the total probability remains consistent across all slices, allowing quantitative comparison of relative peak amplitudes and modal redistribution induced by the penalties.

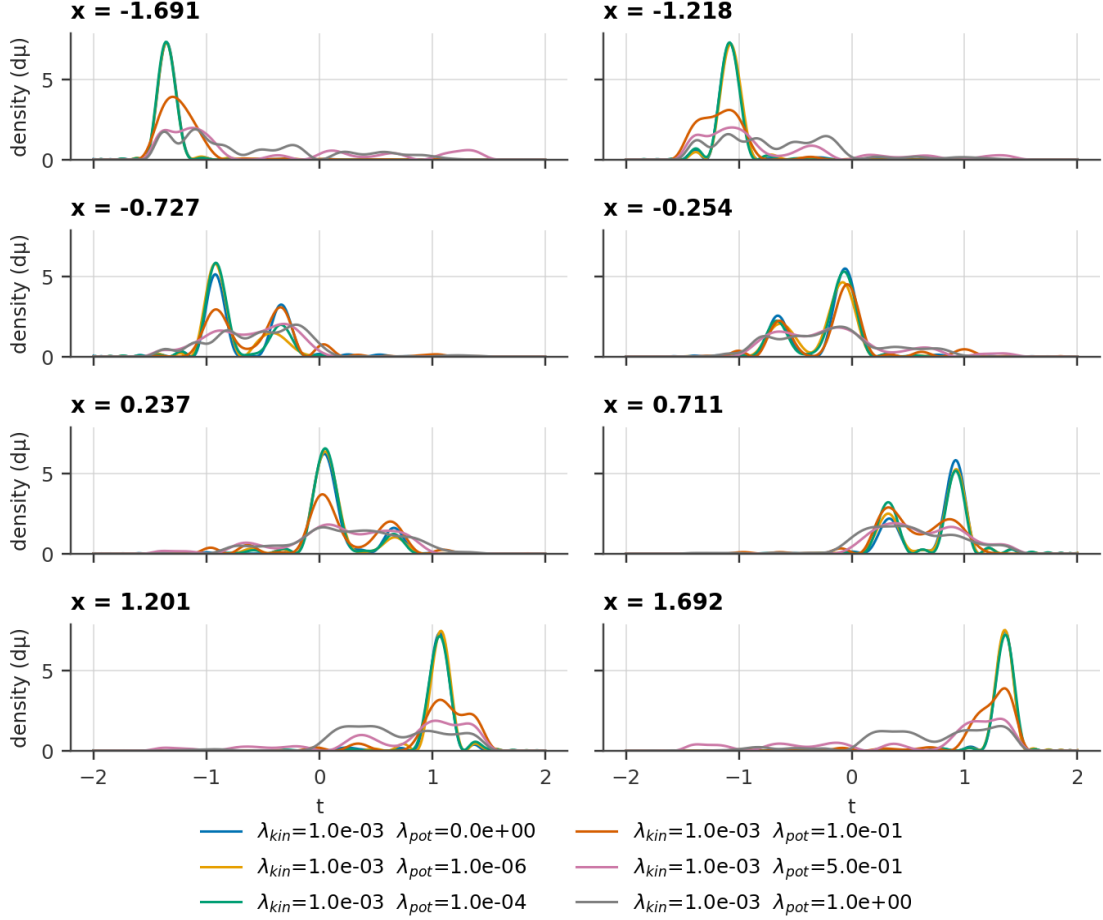

**Figure 10.** Conditional density slices at ten representative $x$-locations for the non-invertible inverse map (21), overlaid across all $\lambda_{\text{pot}} \in \{0, 10^{-6}, 10^{-4}, 0.1, 0.5, 1\}$ values, illustrating the impact of increasing potential regularization (using harmonic oscillator potential) on peak sharpness, modal structure, and redistribution of probability mass.

## 9. Evaluating Multimodal Predictions: A Framework for SNN Verification

Evaluating a SNN in settings where the forward process is non-monotone requires verification at the level of conditional densities rather than point predictions. Let $\mathbf{x} \in \mathcal{X} \subset \mathbb{R}^{d_{\mathbf{x}}}$ denote an input and $y \in \mathcal{Y} \subset \mathbb{R}$ a target; in the inverse problems that motivate the SNN, each $\mathbf{x}$ may admit several plausible $y$-values. The appropriate reference object is therefore the conditional distribution $p^{\star}(y|\mathbf{x})$, which can be sharply multimodal, and the learned object is the SNN-induced conditional density $p^{\star\star}(y|\mathbf{x})$ obtained from a normalized wavefunction. The central question is whether $p^{\star\star}(\cdot|\mathbf{x})$ faithfully reproduces the modal structure and mass allocation of $p^{\star}(\cdot|\mathbf{x})$ across the input domain. To address this question in a manner that is consistent, reproducible, and sensitive to the three principal aspects of multimodal behavior— the number of modes present, the locations of those modes, and the allocation of probability mass among them— we introduce a density-level verification framework with carefully matched measures, grids, and metrics.

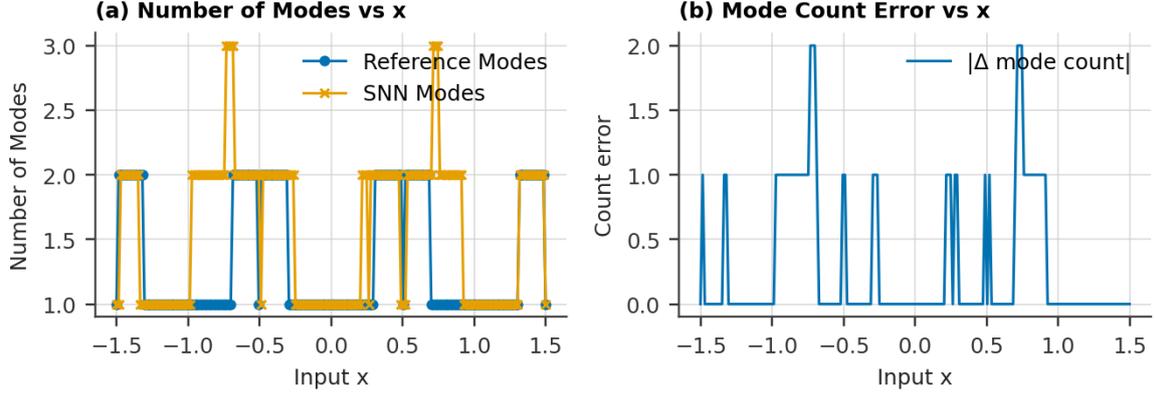

**Figure 11.** (a) Number of modes as a function of input for the non-invertible inverse map (21). For each evaluation input $x$, the number of strict local maxima ("modes") of the reference conditional density $p^\star(y|x)$ and of the SNN-induced density $p^{\star\star}(y|x)$ is computed along the common $y$-grid using the same Chebyshev measure, with peak detection based on prominence thresholds of $10^{-4}$ (reference) and 0.4 (SNN). (b) Absolute mode-count error versus input. The curve shows $E_{\text{count}}(x)$ computed at each input using the same mode-finding procedure and prominence thresholds. Peaks in this curve directly mark inputs where modality is mismatched.

Constructing the reference conditionals proceeds from the known forward model $\mathbf{x} = f(y) + \epsilon$, where $y \sim p_Y$, $\epsilon \sim p_\epsilon$ with $y$ independent of $\epsilon$ and $\epsilon$ is noise with known law. All normalizations and integrals below are taken with respect to the same measure $\mu$ on the $y$–axis that the SNN employs for orthonormality (e.g., $d\mu(y) = w(y)dy$ for a Chebyshev weight). When the forward model $f$ and noise law $p_\epsilon$ are known, Bayes' rule yields

$$p^\star(y|\mathbf{x}) \propto p_\epsilon(\mathbf{x} - f(y))p_Y(y), \tag{43}$$

with the prior $p_Y$ induced by the generative process. When a closed form is unavailable or intractable, a high-fidelity Monte Carlo approximation suffices: draw $y^{(m)} \sim p_Y$, $\epsilon^{(m)} \sim p_\epsilon$ form $\mathbf{x}^{(m)} = f(y^{(m)}) + \epsilon^{(m)}$, and for each $\mathbf{x}$ of interest approximate the conditional by stratification (kernel conditioning) on the set of indices whose simulated inputs are close to $\mathbf{x}$. In all cases, one obtains a matrix $\mathbf{P}^\star \in \mathbb{R}_+^{N_y \times N_\mathbf{x}}$ of column-normalized densities,

$$\mathbf{P}^\star_{i,j} \approx p^\star(y_i|\mathbf{x}_j), \quad \sum_{i=1}^{N_y} \mathbf{P}^\star_{i,j}\, \omega_i = 1, \quad j = 1, \ldots, N_\mathbf{x}, \tag{44}$$

on the same grid $\{y_i\}$ and with the same weights $\{\omega_i\}$ used to evaluate the SNN heat map $\mathbf{P}^{\star\star}_{i,j} \approx p^{\star\star}(y_i|\mathbf{x}_j)$.

With reference conditionals in hand, the core task is to compare modal structure across inputs. For each input column, extract local maxima from the SNN heat map and from the reference density using a consistent peak-finding rule that includes a prominence threshold $\rho > 0$ to suppress spurious ripples. This yields, for each input, a set of predicted mode positions and a set of true mode positions. For a discrete column $p_i = \mathbf{P}_{i,j}$, a grid index $i^\star$ is retained as a mode if $p_{i^\star}$ exceeds its nearest bracketing minima by at least $\rho$ and if the discrete second derivative is negative. Denoting by $\mathcal{M}^\star(\mathbf{x}_j)$ and $\mathcal{M}^{\star\star}(\mathbf{x}_j)$ the sets of indices of retained modes for the reference and SNN densities, respectively, the per-input mode count error is

$$E_{\text{count}}(\mathbf{x}_j) = \left| |\mathcal{M}^\star(\mathbf{x}_j)| - |\mathcal{M}^{\star\star}(\mathbf{x}_j)| \right|. \tag{45}$$

Aggregating over $j$ yields a profile of missed and spurious modes, which is especially informative in regions where the map $f$ folds and the inverse $p^\star(\cdot|\mathbf{x})$ changes its modality.

Figure 11(a) plots $|\mathcal{M}^\star(x_j)|$ and $|\mathcal{M}^{\star\star}(x_j)|$ across the input domain for the inverse problem (21), thereby exposing agreements and disagreements in predicted modality. Agreement between the two curves indicates that the SNN recovers the correct modal count of the inverse map at that input, a necessary precondition for faithful multimodal inference. Deviations mark inputs where $p^{\star\star}(\cdot|x)$ either suppresses a true branch (under-count) or hallucinates a spurious peak (over-count). In typical inverse problems, discrepancies cluster near fold regions—where the forward map locally flattens and $p^\star(\cdot|x)$ changes modality—making this plot an effective locator of ill-posed zones. Because both densities are evaluated on the same grid and normalized under the same measure, observed differences reflect modeling behavior rather than numerical artifacts. Sustained under-counts suggest overly strong smoothness or insufficient basis order, whereas oscillatory over-counts often indicate residual ringing in the learned density. Figure 11(b) reveals where, and how severely, the SNN disagrees with the ground-truth modal structure. Local spikes typically correspond to branch-creation or -annihilation events in the reference density that the SNN fails to mirror, or to small-amplitude ripples in $p^{\star\star}$ that clear the prominence threshold and are misidentified as modes. Because the error is integer-valued, aggregations such as its mean and upper quantiles are robust to outliers and provide a clear target for model selection and hyperparameter tuning. In practice, reducing $E_{\text{count}}(x)$ is best achieved by (i) modestly increasing basis order, (ii) calibrating the regularization to suppress short-wavelength oscillations without erasing weak but genuine

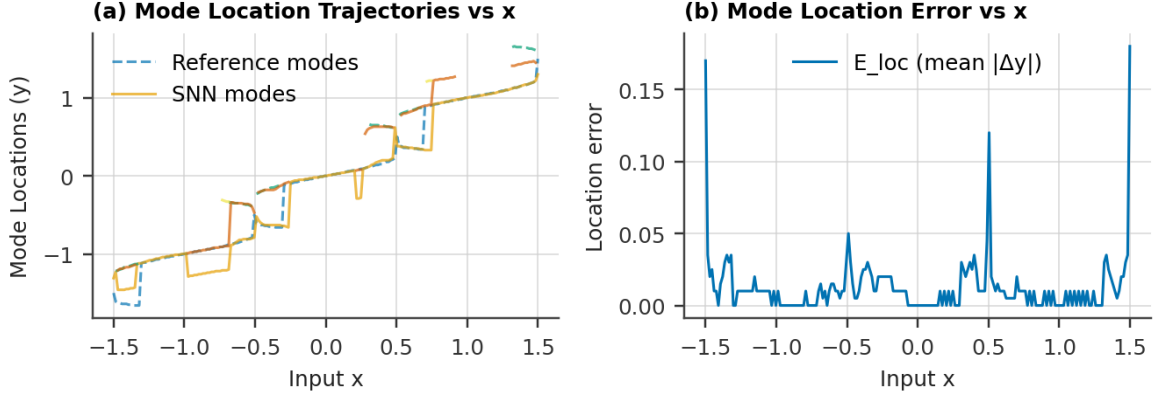

**Figure 12.** (a) Modal location trajectories for the non-invertible inverse map (21). Each curve traces the $y$-location of a detected density mode as the input $x$ varies. For every $x$, modes are identified independently in (i) the analytic reference conditional $p^\star(y|x)$ and (ii) the SNN estimate $p^{\star\star}(y|x)$ by peak finding with prominences (reference $10^{-4}$, SNN 0.4). To assemble trajectories, the $j$-th most prominent peak at each $x$ is placed in column $j$; missing peaks are padded with NaN, which renders visible gaps where a branch (mode) disappears or reappears. Thus, each branch (column $j$) is one continuous "mode track" whenever the corresponding peak persists across $x$. Dashed lines show reference mode branches; solid lines show SNN mode branches. Colors cycle per plotted line; with two branches this yields four distinct colors (reference+SNN for branch 0, reference+SNN for branch 1). Axes: horizontal—input $x$; vertical—mode location $y$. Interpretation: agreement between dashed and solid curves on the same branch indicates that the SNN recovers the correct mode geometry (positions and bifurcations) across $x$. Divergence between the paired curves signals location errors even when the mode count matches. Gaps or branch splitting reflect topological changes in the conditional (mode birth/merger/vanishing) rather than plotting artifacts. (b) Mean absolute mode-location error versus input. For each $x$, true and predicted modes are matched by solving a minimal-cost assignment in physical $y$-space; the per-input location error is $E_{\mathrm{loc}}(x)$. Unmatched modes are penalized by an additive term scaled by the local error level. Modes were detected using the same prominence thresholds.

branches, and (iii) confirming that the chosen prominence values remain fixed across experiments.

Beyond mere presence or absence, the fidelity of mode locations must be quantified. Because mode identities can swap as $\mathbf{x}$ varies, direct indexwise comparison is ill-posed; instead we match predicted and true modes via an assignment that minimizes a physically meaningful cost. Let $y_1^\star(\mathbf{x}_j), \dots, y_{m_j^\star}^\star(\mathbf{x}_j)$ be the true modal locations and $y_1^{\star\star}(\mathbf{x}_j), \dots, y_{m_j^{\star\star}}^{\star\star}(\mathbf{x}_j)$ the predicted ones, with $m_j^\star = |\mathcal{M}^\star(\mathbf{x}_j)|$ and $m_j^{\star\star} = |\mathcal{M}^{\star\star}(\mathbf{x}_j)|$. Define the cost matrix

$$C_{rs}(\mathbf{x}_j) = \frac{|y_r^{\star\star}(\mathbf{x}_j) - y_s^\star(\mathbf{x}_j)|}{s_j} + \lambda_\kappa |\kappa_r^{\star\star}(\mathbf{x}_j) - \kappa_s^\star(\mathbf{x}_j)|, \tag{46}$$

where $s_j$ is a local scale (e.g., inter-modal spacing or inter-quartile range), $\kappa$ denotes negative curvature estimated by a second-order finite difference at the mode, and $\lambda_\kappa \geq 0$ moderates the curvature penalty. Solving the bipartite assignment with the Hungarian algorithm [42, 43] yields a partial matching $\pi_j \subset \{1, \dots, m_j^{\star\star}\} \times \{1, \dots, m_j^\star\}$ that minimizes total cost. Unmatched predictions are counted as false positives; unmatched true modes are misses. The location error at $\mathbf{x}_j$ is then summarized by

$$E_{\mathrm{loc}}(\mathbf{x}_j) = \frac{1}{|\pi_j|} \sum_{(r,s) \in \pi_j} |y_r^{\star\star}(\mathbf{x}_j) - y_s^\star(\mathbf{x}_j)|, \tag{47}$$

and its distribution is reported via mean, median, and high-quantile (e.g., 95th percentile) summaries along $j$. In practice, large $E_{\mathrm{loc}}$ values concentrate in "bifurcation" zones where $\partial f / \partial t$ is small and $p^\star(\cdot|\mathbf{x})$ is rapidly varying. Visualizing the trajectories of mode locations as functions of input, for both the SNN and the reference, is informative in these regions. Disagreement in these "bifurcation" zones typically indicates either insufficient basis resolution or overly aggressive smoothness regularization in the SNN. Tracking these mode curves also helps detect mode crossing and merging events, which are inevitable in multimodal problems and must be handled carefully to avoid artificial label switching in diagnostics.

Figure 12(a) represent modal location trajectories for the inverse problem (21). The trajectory overlay tests not only whether modes exist but also whether they sit in the right places as $x$ varies. Tight co-alignment of reference and SNN ridges across the domain indicates accurate recovery of branch geometry, including the correct handling of merges and splits. Systematic lateral offsets reveal location bias, while broken SNN trajectories that skip branches point to missed modes; conversely, short stray SNN segments unaccompanied by reference ridges indicate spurious peaks. Because the same prominence thresholds govern both extractions, differences primarily reflect modeling rather than extraction bias. This figure should be read in tandem with the count and allocation plots: misaligned trajectories that co-occur with mass misallocation typically signal that the model attributes probability to the wrong branch even when the overall uncertainty is well calibrated. Figure 12(b) quantifies how far predicted modes drift from their true locations independent of mode count. Elevated errors localized to narrow input intervals typically coincide with rapid geometric changes in the reference ridges (e.g., near bifurcations), where small errors in the learned coefficients produce large shifts in modal position. Broad plateaus of elevated error can indicate underfitting (basis too low) or excessive smoothing that blunts curvature at peaks, whereas sharp, isolated spikes often trace back to brief mode-count discrepancies. Because the assignment is performed in $y$-space and normalized per input, the measure is comparable across the domain and is directly actionable for tuning basis order and curvature-aware regularization.

Mode mass is the third pillar of comparison. For each input, one can define a basin of attraction around each true mode by assigning target grid points to the nearest true mode location, or more robustly by a watershed segmentation on the negative density landscape. Integrating the density over each basin yields a probability allocation per mode. Applying the same partition to the SNN density produces predicted allocations. The difference between predicted and true allocations, summarized across inputs, reveals whether the SNN overconcentrates mass on dominant modes, underrepresents minor branches, or artificially redistributes mass in ambiguous zones. This basin-based view is more stable than comparing raw peak heights because it reflects the entire local structure around a mode rather than a single grid point. For a fixed input $\mathbf{x}_j$, define a partition of $\mathcal{Y}$ into basins of attraction around the true modal locations. A simple and robust partition is the Voronoi tessellation [44-46] induced by $\{y_s^\star(\mathbf{x}_j)\}_{s=1}^{m_j^\star}$: for each grid point $y_i$, assign it to the nearest true mode by Euclidean distance along $y$. Denote the basin of mode $s$ by $B_s(\mathbf{x}_j) \subset \{1,\ldots,N_y\}$. The reference allocation vector $\alpha^\star(\mathbf{x}_j)$ has components

$$\alpha^\star(\mathbf{x}_j) = \sum_{i \in B_s(\mathbf{x}_j)} \mathbf{P}_{i,j}^\star \omega_i, \qquad s = 1, \ldots, m_j^\star, \tag{48}$$

and the SNN allocation $\alpha^{\star\star}(\mathbf{x}_j)$ is obtained by integrating $\mathbf{P}_{\cdot,j}^{\star\star}$ over the same partition:

$$\alpha^{\star\star}(\mathbf{x}_j) = \sum_{i \in B_s(\mathbf{x}_j)} \mathbf{P}_{i,j}^{\star\star} \omega_i. \tag{49}$$

Because both use the same weights $\{\omega_i\}$ and basins $B_s(\mathbf{x}_j)$, the difference $\alpha^{\star\star}(\mathbf{x}_j) - \alpha^\star(\mathbf{x}_j)$ measures redistribution of probability mass across true modes. We summarize allocation fidelity via an $\ell_1$ discrepancy

$$E_{\text{alloc}}(\mathbf{x}_j) = \frac{1}{2}\|\alpha^{\star\star}(\mathbf{x}_j) - \alpha^\star(\mathbf{x}_j)\|_1 = \frac{1}{2}\sum_{s=1}^{m_j^\star} |\alpha_s^{\star\star}(\mathbf{x}_j) - \alpha_s^\star(\mathbf{x}_j)|, \tag{50}$$

which lies in [0,1] and is insensitive to local oscillations within basins. When the negative density landscape is rugged, a watershed segmentation [47, 48] on $-\mathbf{P}_{\cdot,j}^\star$ can replace the Voronoi rule; in our experience, both lead to consistent conclusions provided the segmentation is applied identically to $\mathbf{P}^\star$ and $\mathbf{P}^{\star\star}$.

Figure 13(a) represents mode mass allocation at representative inputs for the inverse problem (21). Basin-level allocation offers a stable view of multimodality that is less sensitive to peak sharpness than raw peak heights. Visual agreement between the stacked bars indicates that the SNN distributes probability mass among the correct branches, reflecting fidelity in practical decision-making tasks (e.g., selecting among multiple plausible solutions). Under-allocation to weaker branches and over-allocation to dominant ones reveal overconfidence and smoothing bias; the inverse pattern suggests spurious shoulder peaks siphoning probability away from the main mode. Because the partition is always defined with respect to the reference modes, differences in the stacks can be interpreted unambiguously as mass redistribution errors rather than geometric artifacts of the learned density. Figure 13(b) summarizes, at each input, how much probability mass the SNN shifts across true branches. Low values confirm that the model concentrates uncertainty where it should, even if peaks differ slightly in height or sharpness; high values indicate substantive misallocation, such as neglecting a minor yet genuine branch or assigning mass to an absent mode. In practice, peaks in reports $E_{\text{alloc}}(\mathbf{x})$ often co-occur with spikes in mode-count or location error but can also arise in isolation when the SNN captures the geometry yet miscalibrates relative masses. As a consequence, this curve is a sensitive indicator of calibration quality.

Global distributional diagnostics complement these modal metrics and characterize accuracy beyond modes. The entropy of the conditional as a function of input [49],

$$H^\star(\mathbf{x}) = -\int_{\mathcal{Y}} p^\star(y|\mathbf{x}) \log p^\star(y|\mathbf{x}) \, d\mu(y), \qquad H^{\star\star}(\mathbf{x}) = -\int_{\mathcal{Y}} p^{\star\star}(y|\mathbf{x}) \log p^{\star\star}(y|\mathbf{x}) \, d\mu(y), \tag{51}$$

highlights regions where the inverse is intrinsically diffuse; a well-calibrated SNN should track the shape of $H^\star(\mathbf{x})$. Discrepancies in entropy profiles can reveal over-smoothing (excess entropy) or over-confidence (deficit entropy). To localize mismatches, we compute a symmetric divergence along the input axis, such as the Jensen-Shannon (JS) distance [50-52],

$$\text{JS}(p^\star(\cdot|\mathbf{x}) \| p^{\star\star}(\cdot|\mathbf{x})) = \frac{1}{2}\text{KL}(p^\star \| m) + \frac{1}{2}\text{KL}(p^{\star\star} \| m), \qquad m = \frac{1}{2}(p^\star + p^{\star\star}), \tag{52}$$

with each KL divergence evaluated under $\mu$. In addition, ridge maps—the zero-contours of the derivative $\partial p(y|\mathbf{x})/\partial y$ with negative second derivative—trace modal trajectories as functions of $\mathbf{x}$; plotting the ridges of $p^\star$ and $p^{\star\star}$ on the same axes makes crossing and merging events visually salient. Finally, we use highest posterior density (HPD) credible sets, which are less sensitive to peak sharpness. For any credibility level $\gamma \in (0,1)$ and fixed input $\mathbf{x}$, the HPD set of the reference conditional is the density level set at a threshold $\tau_\gamma^\star(\mathbf{x})$ chosen to enclose exactly $\gamma$ posterior mass:

$$S_\gamma^\star(\mathbf{x}) = \{y: p^\star(y|\mathbf{x}) \geq \tau_\gamma^\star(\mathbf{x})\}, \quad \int_{S_\gamma^\star(\mathbf{x})} p^\star(y|\mathbf{x}) d\mu(y) = \gamma. \tag{53}$$

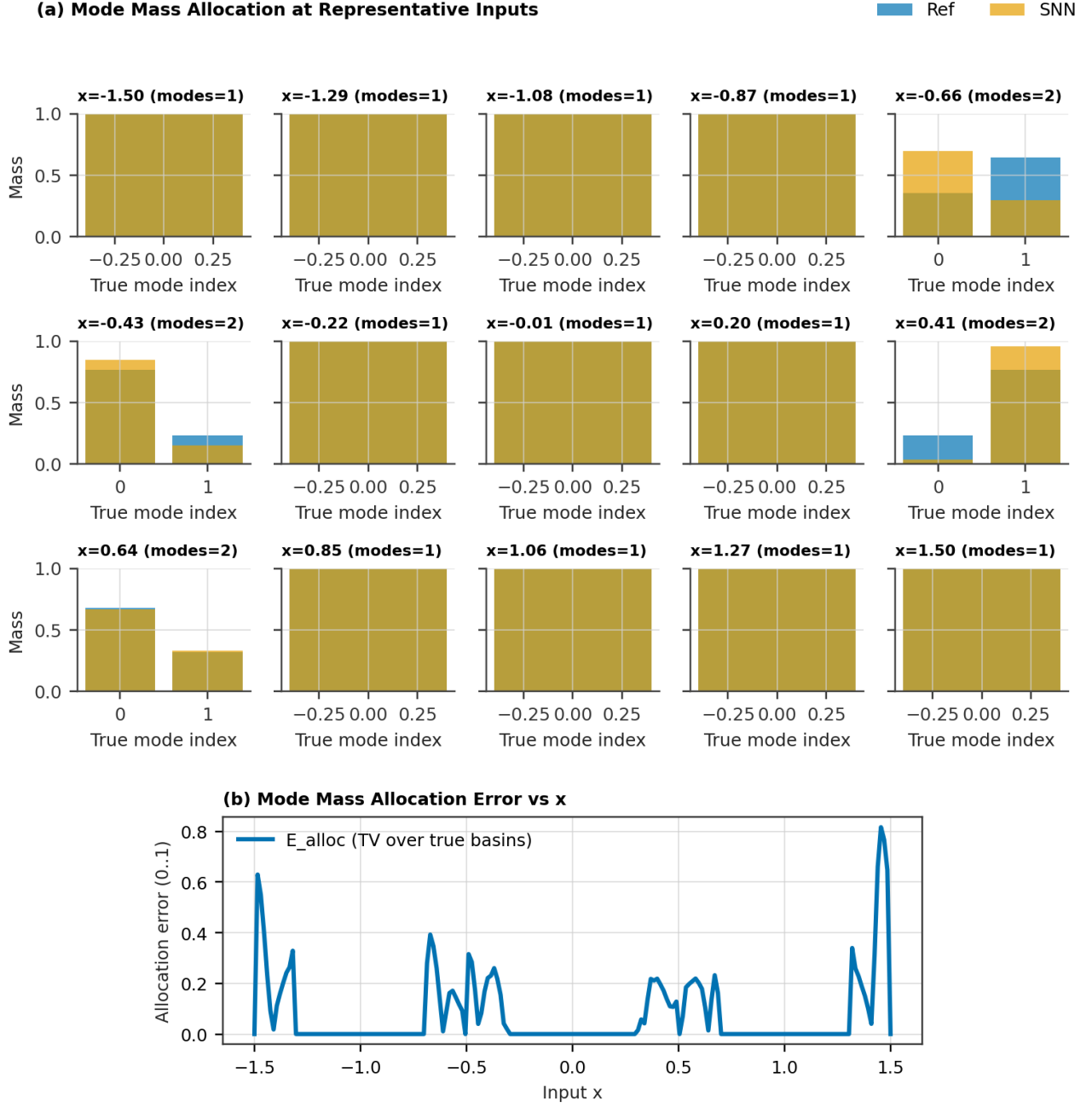

**Figure 13.** (a) Mode mass allocation at representative inputs for the non-invertible inverse map (21). At several inputs $x$, the output domain is partitioned into Voronoi basins induced by the true modal locations. The bars show the mass allocated by the reference density $\alpha^\star(x)$ and by the SNN $\alpha^{\star\star}(x)$ to each true basin, computed under the common Chebyshev measure and normalized to sum to one. Mode sets are defined with the same prominence thresholds (reference $10^{-4}$, SNN 0.4) used elsewhere. (b) Mode mass allocation error versus input. The curve reports $E_{\text{allloc}}(x) \in [0,1]$, the total variation distance between the SNN and reference allocation vectors over the true Voronoi basins.

Define $S_\gamma^{\star\star}(\mathbf{x})$ analogously for $p^{\star\star}(y|\mathbf{x})$. To quantify how similarly the two posteriors localize their $\gamma$-credible regions, we compute the Jaccard index (with respect to $\mu$) [53-55]:

$$J_\gamma(\mathbf{x}) = \frac{\mu\left(S_\gamma^\star(\mathbf{x}) \cap S_\gamma^{\star\star}(\mathbf{x})\right)}{\mu\left(S_\gamma^\star(\mathbf{x}) \cup S_\gamma^{\star\star}(\mathbf{x})\right)} \in [0,1], \tag{54}$$

where larger values indicate greater geometric overlap for the same mass budget $\gamma$.

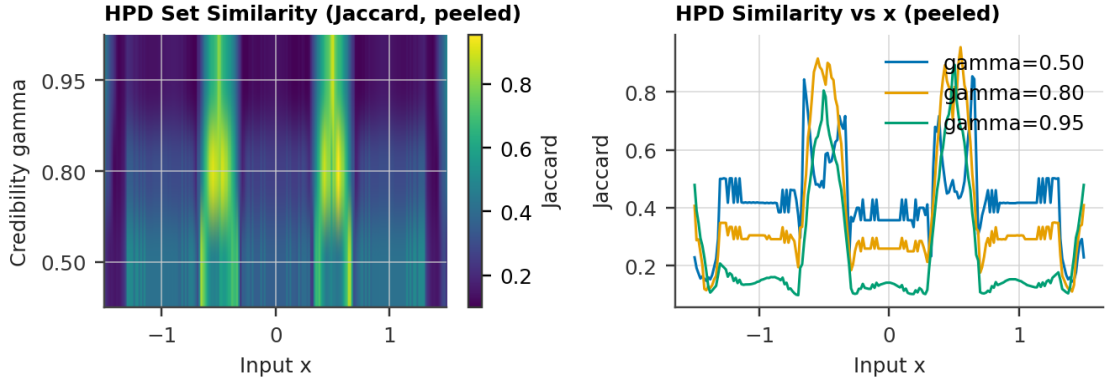

**Figure 14.** Endpoint-aware HPD set similarity between the learned conditional $p^{**}(y|x)$ and the analytic reference $p^*(y|x)$ for the non-invertible inverse map (21). Left: weighted Jaccard heatmap $J_\gamma(x)$ across credibility levels $\gamma \in \{0.50, 0.80, 0.95\}$ after peeling the outer 5% of the $y$-grid at both ends and renormalizing on the interior under Chebyshev weights. Right: Jaccard profiles $J_\gamma(\cdot)$ vs. $x$ for each $\gamma$. Larger values indicate closer agreement of HPD level sets; localized dips mark mass-allocation mismatches or spurious/missing modes.

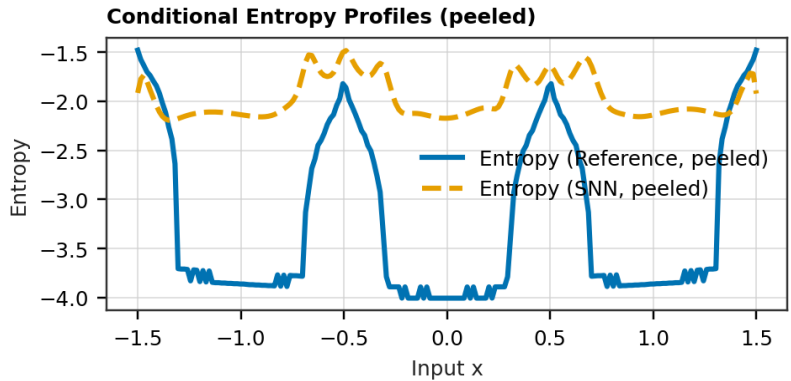

**Figure 15.** Endpoint-aware conditional entropy profiles $H(x)$ for the reference (solid) and SNN (dashed), for the non-invertible inverse map (21), computed on the peeled interior using Chebyshev weights after removing the outer 5% of the $y$-grid, applying light Savitzky–Golay smoothing, and renormalizing each row. Agreement between curves indicates well-calibrated dispersion of $p^*$ relative to $p^{**}$; higher SNN entropy signals over-smoothing, whereas lower entropy indicates over-confidence or mode collapse.

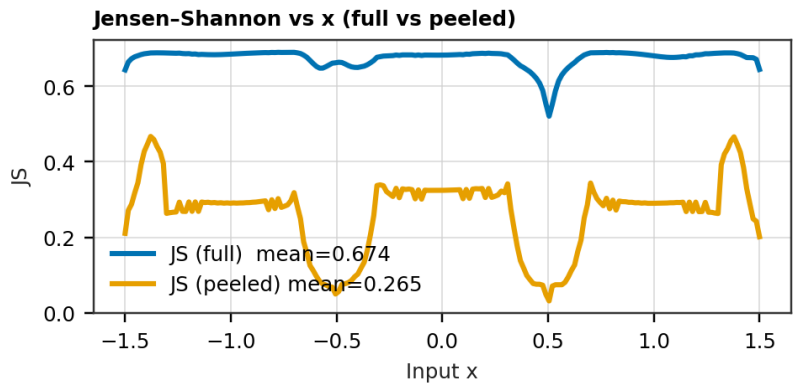

**Figure 16.** JS divergence between $p^*(y|x)$ and $p^{**}(y|x)$ as a function of $x$, for the non-invertible inverse map (21), comparing evaluation on the full $y$-grid versus the peeled interior (outer 5% removed, rows gently smoothed and renormalized). JS is computed with Chebyshev-1 quadrature weights; 0 denotes identity and $\log 2$ the maximal discrepancy. The peeled curve discounts boundary instabilities inherent to orthogonal-polynomial representations and emphasizes interior fidelity.

For the inverse problem (21), on the peeled interior, shape agreement of credible regions is further supported by high weighted Jaccard similarity across $\gamma \in \{0.50, 0.80, 0.95\}$, with localized dips flagging $x$-intervals of mass misallocation or missing modes (see figure 14). The SNN's conditional entropies track the analytic reference, and the JS profile indicates that most discrepancies observed on the full grid are boundary-driven and largely vanish after peeling (see figures 15 and 16).

Regularization exerts a first-order influence on the three axes of evaluation and should be tuned with these diagnostics in mind. A smoothness penalty on the wavefunction or the density, implemented as a quadratic form in coefficients, reduces spurious peaks and stabilizes mode trajectories but can blur genuine secondary modes if set too high, manifesting as depressed mode counts and inflated allocation errors against minor branches. Conversely, a potential term that increases away from physically plausible target regions can suppress boundary artifacts and guide mass away from impossible states, but may distort allocations if it conflicts with the true structure. The trade-offs are best visualized by sweeping regularization strengths and plotting $E_{\text{count}}$, $E_{\text{loc}}$ quantiles, and $E_{\text{alloc}}$ across $\mathbf{x}$. In many problems, a moderate smoothness level combined with a weak, well-targeted potential achieves the best balance between faithful multimodality and numerical stability.

We now formalize the principal metrics used throughout. For a set of inputs $\mathbf{x}_1, \ldots \mathbf{x}_{N_{\mathbf{x}}}$, define the mean mode count error and its tail summary as

$$\bar{E}_{\text{count}} = \frac{1}{N_{\mathbf{x}}} \sum_{j=1}^{N_{\mathbf{x}}} E_{\text{count}}(\mathbf{x}_j), \qquad Q_{\text{count}}^{(q)} = \text{Qunatile}_q\left(\{E_{\text{count}}(\mathbf{x}_j)\}_j\right), \tag{55}$$

and for locations, using matchings $\pi_j$,

$$\bar{E}_{\text{loc}} = \frac{1}{N_{\mathbf{x}}} \sum_{j=1}^{N_{\mathbf{x}}} E_{\text{loc}}(\mathbf{x}_j), \qquad Q_{\text{loc}}^{(q)} = \text{Qunatile}_q\left(\{E_{\text{loc}}(\mathbf{x}_j)\}_j\right). \tag{56}$$

Allocation accuracy is summarized by

$$\bar{E}_{\text{alloc}} = \frac{1}{N_{\mathbf{x}}} \sum_{j=1}^{N_{\mathbf{x}}} E_{\text{alloc}}(\mathbf{x}_j), \qquad Q_{\text{alloc}}^{(q)} = \text{Qunatile}_q\left(\{E_{\text{alloc}}(\mathbf{x}_j)\}_j\right). \tag{57}$$

For global geometry,

$$\bar{H}^{\star} = \frac{1}{N_{\mathbf{x}}} \sum_{j=1}^{N_{\mathbf{x}}} H^{\star}(\mathbf{x}_j), \qquad \bar{H}^{\star\star} = \frac{1}{N_{\mathbf{x}}} \sum_{j=1}^{N_{\mathbf{x}}} H^{\star\star}(\mathbf{x}_j), \qquad \overline{\text{JS}} = \frac{1}{N_{\mathbf{x}}} \sum_{j=1}^{N_{\mathbf{x}}} \text{JS}(p^{\star}(\cdot|\mathbf{x}_j) \| p^{\star\star}(\cdot|\mathbf{x}_j)), \tag{58}$$

and for HPD set similarity at credibility $\gamma$,

$$\bar{J}_\gamma = \frac{1}{N_{\mathbf{x}}} \sum_{j=1}^{N_{\mathbf{x}}} J_\gamma(\mathbf{x}_j). \tag{59}$$

Each of these quantities is computed with quadrature weights $\{\omega_i\}$, ensuring that the comparisons are conducted under the same measure that defines SNN orthogonality and normalization.

## 10. Multivariate Outputs

While the base construction treats scalar outputs, multivariate outputs $\mathbf{y} \in \mathbb{R}^m$ are handled by tensor-product bases $\{\prod_{j=1}^m \varphi_{k_j}(\xi_j)\}$ on a hyperrectangle mapped to $[-1,1]^m$. The coefficient tensor $C(x)$ then has $(K+1)^m$ entries, which is infeasible for large $m$. Two scalable strategies mitigate this curse of dimensionality. First, use separable expansions of bounded rank,

$$\psi_x(\boldsymbol{\xi}) \approx \sum_{r=1}^{R} \prod_{j=1}^{m} \sum_{k=0}^{K} a_{jrk}(x) \varphi_k(\xi_j), \tag{60}$$

(which is a canonical polyadic (CP) / rank-$R$ factorization of the $(K+1)^m$ tensor of coefficients.) so the parameter count scales as $O(R\,m\,K)$ and can be generated by the network in block-structured heads. Second, adopt a mixture-of-separables representation,

$$\psi_x = \sum_{r=1}^{R} \alpha_r(x) \psi_{xr}^{(1)}(\xi_1) \ldots \psi_{xr}^{(m)}(\xi_m), \tag{61}$$

with mixing weights $\alpha_r(x)$. Because the univariate bases are orthonormal, inner products and the global $L^2$ norm admit closed-form expressions via contractions of small Gram objects derived from the univariate factors, so normalization can be enforced analytically without materializing the full tensor. These constructions capture cross-coordinate dependence via superposition while avoiding exponential blow-up. Complete derivations, identifiability and rank-selection criteria, conditioning and normalization schemes, and training/complexity analyses will be presented in our next paper.

## 11. Conclusions and Discussion

We introduced the SNN as an amplitude-first framework for conditional density estimation that reconciles expressive modeling of multimodality with exact likelihoods and analytic downstream quantities. By mapping each input to a normalized spectral amplitude and reading probabilities via Born's rule, the SNN elevates probabilistic prediction from parametrized likelihood heads to a physically inspired representation in which positivity and normalization are structural rather than enforced. This design yields three practical strengths: (i) densities are nonnegative and exactly normalized by construction; (ii) complex, multimodal, and asymmetric conditionals emerge naturally through interference among basis modes without mixture bookkeeping or invertibility constraints; and (iii) moments, quantiles, calibration diagnostics, and constraint evaluations reduce to precomputable quadratic forms in coefficient space, enabling fast, reproducible analysis.

Methodologically, the paper developed the statistical and computational foundations that make this representation practical. First, exact maximum-likelihood training is achieved by parameterizing coefficient vectors on the unit sphere, aligning optimization with the geometry that guarantees normalization. Second, physics-inspired quadratic regularizers—kinetic energy to penalize spectral roughness and soft confining potentials to manage tail mass—provide transparent, tunable control of uncertainty concentration versus complexity, clarifying the trade-offs that typically remain heuristic in alternative approaches. Third, an operator calculus closes the loop between modeling and supervision: observables, constraints, and weak labels are expressed as self-adjoint matrices acting on amplitudes, so that safety margins, partial information, and measurement rules enter training as cheap quadratic forms. Finally, scalable extensions for multivariate outputs via separable and low-rank tensor constructions preserve the core SNN advantages while mitigating the curse of dimensionality.

Positioned against prominent baselines, the SNN combines desirable features that are rarely available in a single method. Like MDNs, it captures multimodality, but without choosing a component count or suffering label switching; like NFs, it offers exact likelihoods, but without the architectural and Jacobian constraints; like EBMs, it supports flexible shapes, but training requires no negative sampling and no partition-function estimation; and unlike quantile or conformal methods, it yields a full generative law from which arbitrary risk functionals can be computed analytically or with high-accuracy quadrature. Equally important, the spectral parameterization exposes intelligible "dials"—basis order, compactification map, and phase allowance—that connect directly to approximation theory and numerical stability, improving interpretability and capacity control.

The amplitude perspective also brings novel diagnostics. Because entropy, kinetic energy, nodal structure, and other geometry-of-amplitude quantities are quadratic or low-order forms in coefficient space, calibration checks, OOD cues, and smoothness measures can be computed cheaply and reproducibly. This facilitates principled model selection and ablation: one can, for example, read off whether excess entropy signals oversmoothing, whether elevated kinetic energy indicates spurious oscillations, or whether nodal instability under perturbations reveals brittle multimodal fits.

Limitations suggest clear paths for improvement. Truncated spectral expansions require careful choices of basis order and domain mapping; poor compactification or under-resolved spectra can induce boundary artifacts or spurious oscillations. Although separable and low-rank constructions substantially mitigate the growth of parameters with output dimension, very high-dimensional targets still pose challenges that may require tensor-network factorizations, multiresolution bases, or convolutional spectral layers. While complex coefficients are essential to interference and expressivity, they shift interpretability from parameter space to operator action and diagnostic geometry; practitioners may need tooling to visualize amplitude phases, interference patterns, and their effect on predictive structure.

These limitations motivate several concrete research directions:
1. Adaptive representations. Learn compactification maps and basis order end-to-end, explore multiresolution (wavelet) or piecewise-orthogonal bases, and develop data-driven kinetic/potential schedules that adapt to local complexity and tail behavior.
2. High-dimensional structure. Combine SNN amplitudes with tensor decompositions, convolutional spectral layers for images and fields, and graphical or operator-factorized constructions to capture conditional dependencies without full tensor products.
3. Hybrid models. Couple amplitudes with flow-based transports or score models to inherit both analytic quadratic calculus and scalable high-dimensional expressivity; explore Bayesian SNNs to propagate parameter uncertainty into amplitude geometry.
4. Riemannian optimization. Exploit the unit-sphere (or Stiefel) geometry for stable training, with trust-region and natural-gradient variants that respect normalization constraints while accelerating convergence.
5. Generalized supervision. Extend the operator framework to positive operator-valued measures for weak labels, interval/censored outcomes, fairness or safety constraints, and physics-informed observables—keeping all supervision quadratic and sparse.
6. Decision-theoretic tooling. Package closed-form integrals for common losses (pinball, Huber, CVaR, Bayes actions) and uncertainty decompositions (entropy, mutual information) to streamline deployment in risk-sensitive applications.
7. Evaluation protocols. Standardize multimodal diagnostics rooted in amplitude geometry (entropy profiles, kinetic spectra, nodal stability) alongside conventional likelihood and calibration metrics, enabling apples-to-apples comparisons with MDNs, NFs, and EBMs.

## Code availability

All code used to generate the data, figures, and analyses in this study is publicly available, together with implementation details, at the Zenodo repository (https://doi.org/10.5281/zenodo.17442982) [56].